\documentclass[preprint,12pt]{elsarticle}

\usepackage[utf8]{inputenc}
\usepackage[english]{babel}
\usepackage{amsmath,amssymb,amsfonts}
\usepackage{booktabs}
\usepackage{graphicx}
\usepackage{hyperref}
\usepackage{tabularx}
\usepackage{multirow}
\usepackage{xcolor}
\usepackage{rotating}
\usepackage{tikz}
\usetikzlibrary{arrows.meta, positioning, matrix, calc}

\begin{document}
\pagestyle{empty}

\begin{frontmatter}

\title{Class-Dependent Hybrid Data Augmentation for Multiclass Migraine Classification under Severe Class Imbalance}

\author{Elvin Som\'on S\'anchez\corref{cor1}}
\ead{elvsomsan@alum.us.es}

\author{Miguel A. Guti\'errez-Naranjo}

\affiliation{organization={Department of Computer Science and Artificial Intelligence \newline University of Seville, Seville, Spain}}

\cortext[cor1]{Corresponding author.}

\begin{abstract}
\textbf{Background and Objective:} Multiclass migraine classification from clinical data is hindered by severe class imbalance, limited sample sizes, and methodological biases such as data leakage and inappropriate evaluation metrics, while the impact of data augmentation design remains insufficiently understood. This study investigates how augmentation strategies affect model reliability and proposes a class-dependent framework for imbalanced clinical classification.

\textbf{Methods:} We conducted a reproducibility-oriented reevaluation of prior migraine classification studies, correcting for data leakage and metric bias, and introduced (i)~a clinically motivated aggregation of two hemiplegic subtypes following ICHD-3~\S1.2.3, (ii)~a class-dependent hybrid augmentation strategy that assigns generators by per-class sample size, and (iii)~the concept of \emph{fidelity asymmetry}, motivating proportionally constrained growth as an alternative to full class balance. Experiments used 400 patients across seven subtypes under a two-stage protocol, evaluated with stratified 5-fold cross-validation and macro-averaged F1.

\textbf{Results:} Correcting these flaws collapsed previously inflated accuracies (up to 99.7\% under leakage) onto a leakage-free macro-F1 scale, on which the best prior model reaches only 0.803 (seven classes). Averaged across the eight classifiers, the framework outperformed individual augmenters (0.862 vs.\ 0.836 Copula, 0.815 CTGAN, 0.801 no augmentation), peaking at 0.914~$\pm$~0.047 (FT-Transformer, proportional augmentation). Clinically motivated class aggregation (FT-Transformer, 0.896~$\pm$~0.038) accounts for most of the gain; the framework's principal contribution is improved average robustness across classifiers.

\textbf{Conclusions:} Augmentation design is a critical and underexplored factor in medical machine learning. Class-dependent and proportionally constrained augmentation improves reliability under severe imbalance, while label design remains the primary determinant of performance.
\end{abstract}

\end{frontmatter}

\section{Introduction}
\label{sec:introduction}

Migraine is one of the most prevalent and disabling neurological disorders
worldwide, affecting approximately 14\% of the adult population and representing
a leading cause of years lived with disability~\cite{Stovner2022-az}. From a
clinical perspective, migraine is a complex neurovascular disorder involving
multiple pathophysiological mechanisms, including cortical spreading depression,
trigeminovascular activation, and altered sensory processing~\cite{Ashina2021Migraine}.
These mechanisms contribute to substantial heterogeneity in symptom presentation,
disease progression, and treatment response, complicating accurate diagnosis and
subtype differentiation in routine clinical practice. This heterogeneity has
direct implications for treatment selection and prognosis, making reliable
subtype classification a clinically relevant objective. Despite advances in clinical understanding, migraine diagnosis remains challenging due to overlapping symptom profiles across the subtypes defined in ICHD-3, which distinguishes over a dozen migraine variants with partially shared clinical criteria~\cite{IHS2018}. 

In this context, machine learning (ML) approaches have
been increasingly explored to support automated classification and clinical
decision-making. Several studies have reported promising results using clinical,
questionnaire-based, and imaging-derived features~\cite{Petrusic2025-ig,Lee2025-jb,Stubberud2024-ou}.
However, recent reviews consistently highlight substantial methodological
limitations, including poor reproducibility, limited external validation, and
frequent deviations from reporting standards~\cite{Petrusic2025-ig,Lee2025-jb,Stubberud2024-ou,Collins2024-tripodai}.

A central challenge in migraine classification is the reliance on small and
highly imbalanced clinical datasets, particularly in multiclass settings where
rare but clinically relevant subtypes may be represented by only a few samples
\cite{Khan2024Migraine,Reddy2025Migraine}. This imbalance reflects real-world
clinical registries but introduces significant difficulties for model training,
evaluation, and reproducibility~\cite{He_Garcia_2009}. When not properly
addressed, models tend to favour majority classes, leading to inflated
performance estimates that do not reflect true clinical utility.

This issue is compounded by the use of inadequate evaluation metrics.
Accuracy and micro-averaged scores are dominated by majority-class predictions
and can obscure poor performance on minority subtypes. In contrast,
macro-averaged F1 assigns equal importance to all classes and provides a more
reliable assessment under severe imbalance~\cite{Powers2011,Saito2015}. For this
reason, we adopt macro-F1 as the primary evaluation metric, as it better reflects
the need for consistent performance across all clinically relevant subtypes.

Beyond class imbalance, an additional and underexplored limitation lies in the
design of data augmentation strategies. Classical oversampling methods such as
SMOTE, as well as more recent generative approaches, are typically applied
uniformly across all classes~\cite{Chawla2002-smote,Blagus2013,Xu2019-ctgan}.
However, this uniform treatment ignores subtype-specific data characteristics
and may generate implausible samples, amplify dataset biases, and inflate
performance estimates. This effect compounds the distortion introduced by class
imbalance and can lead to overly optimistic conclusions.

In this study, we address these limitations through five interrelated
contributions, summarised below and presented in detail in the remainder
of the paper. We conduct a \textit{reproducibility-oriented reevaluation}
of prior migraine classification
studies~\cite{Reddy2025Migraine,Khan2024Migraine}, quantifying the impact
of augmentation design and evaluation protocols on reported performance.
On this corrected baseline, we propose a class-dependent hybrid
augmentation framework that adaptively assigns augmentation strategies to
each subtype based on its sample size, in contrast to conventional
uniform augmentation. The framework, summarised in Fig.~\ref{fig:framework},
partitions classes by a sample-size threshold, assigns a generator to
each tier, and controls the absolute volume of synthetic data through a
discrete growth mode.

The proposed framework is evaluated on a clinical dataset comprising 400
patients described by 22 features across seven migraine subtypes with highly
skewed class distributions. Although moderate in size, this dataset reflects the
intrinsic imbalance of specialised clinical registries. To ensure robust
evaluation, we adopt a two-stage experimental design, considering both the full
seven-class setting and a refined six-class configuration in which the two
hemiplegic migraine subtypes (sporadic and familial), clinically defined by
ICHD-3~\S1.2.3 as variants of the same syndrome, are merged into a single
class. The clinical and statistical rationale for this aggregation is presented
in Section~\ref{subsec:discussion_aggregation}.

The main contributions of this work are:

\begin{enumerate}
\item A reproducibility-oriented reevaluation of migraine classification studies, identifying data leakage and reevaluating every prior-art classifier under a leakage-free per-classifier protocol; the strongest prior model reaches only macro-F1 0.803, far below the inflated accuracies originally reported (full per-model results in \ref{subsec:appendix_sota}).
\item A systematic benchmark of classifiers and augmentation methods under leakage-free stratified 5-fold cross-validation.
\item A clinically and statistically justified aggregation of the two hemiplegic migraine subtypes (sporadic and familial) into a single ICHD-3-grounded category, which addresses the structural performance ceiling observed in the seven-class formulation.
\item A class-dependent hybrid augmentation framework that adapts generation strategies to per-class data characteristics.
\item The introduction of {\it fidelity asymmetry} and the demonstration that proportional augmentation improves stability in low-data multiclass settings.
\end{enumerate}

The remainder of this paper is structured as follows. Section~\ref{sec:related_work}
reviews related work. Section~\ref{sec:methodology} describes the dataset and
methods. Section~\ref{sec:results} presents the results. Section~\ref{sec:discussion} analyses and discusses the findings, and Section~\ref{sec:conclusions} concludes the paper.
\section{Related Work}
\label{sec:related_work}

\subsection{Machine Learning for Migraine Classification}

The application of machine learning (ML) to migraine diagnosis and subtype
classification has grown substantially in recent years, driven by the increasing
availability of clinical, questionnaire-based, and neuroimaging datasets.
Early work in this domain primarily relied on classical supervised learning
algorithms, including Support Vector Machines (SVM), k-Nearest Neighbors (KNN),
Random Forests (RF), and gradient boosting methods, often combined with feature
selection techniques and synthetic oversampling to mitigate class
imbalance~\cite{Khan2024Migraine,Reddy2025Migraine}. More recently, deep
learning and transformer-based architectures originally proposed for
general-purpose tabular data have been increasingly explored in
biomedical applications, including models such as TabNet, SAINT,
FT-Transformer, and TabPFN, which aim to capture complex feature
interactions and heterogeneous data distributions more effectively than
classical methods~\cite{Arik2021TabNet,Somepalli2021SAINT,Gorishniy2021-fttransformer,Hollmann2022TabPFN}.
A comprehensive survey of deep neural architectures for tabular data by Borisov
et al.~\cite{Borisov2022survey} provides a structured overview of these
developments, categorising them according to their treatment of feature
interactions, attention mechanisms, and regularisation strategies. Despite their
theoretical advantages, large-scale empirical studies suggest that performance
gains over well-tuned classical models are often modest, particularly in
small-data regimes typical of clinical
applications~\cite{Grinsztajn2022,ShwartzZiv2022}.

Reported performance in migraine classification studies is frequently
high, though often difficult to interpret in the absence of class-level
analysis. Two prior studies using the same clinical dataset~\cite{Khan2024Migraine,Reddy2025Migraine}
report accuracies between 88\% and 99.7\% under SMOTE-based augmentation,
figures that, as we demonstrate in Section~\ref{subsec:results_exploration},
are substantially inflated by data leakage and the use of
majority-dominated metrics.

However, systematic
reviews consistently caution that such results must be interpreted with care,
as the literature is characterised by substantial methodological heterogeneity,
limited reproducibility, and insufficient external
validation~\cite{Petrusic2025-ig,Lee2025-jb,Stubberud2024-ou}. These concerns
raise fundamental questions regarding the true generalisation performance and
clinical utility of existing models. In response, Petru\v{s}i\'c et
al.~\cite{Petrusic2024-recommendations} recently proposed an explicit
recommendation framework for ML studies in headache classification, emphasising
rigorous evaluation procedures, imbalance-sensitive metrics, and reproducible
reporting as minimum methodological standards. The present work is directly
aligned with these recommendations.

\subsection{Class Imbalance and Evaluation in Medical ML}

Class imbalance is a well-established challenge in supervised learning and is
particularly acute in medical applications, where rare but clinically relevant
conditions are inherently underrepresented~\cite{He_Garcia_2009}. In multiclass
settings, imbalance introduces complex interactions between class frequency,
model capacity, and decision boundaries, often leading to biased models that
favour majority classes.

The use of accuracy as a primary evaluation metric in such contexts has been
widely criticised, as it is dominated by majority-class predictions and may
substantially overestimate model
effectiveness~\cite{Mosquera2024-lj,Hellin2024-qy}. Alternative metrics, such
as macro-averaged F1, have been recommended as more appropriate for imbalanced
multiclass problems, as they assign equal weight to each class regardless of
frequency and better reflect performance on minority
classes~\cite{Powers2011,Saito2015,Sokolova2009}.

Despite these recommendations, many studies in migraine classification continue
to report accuracy as their main outcome, often without class-level performance
analysis. Furthermore, recent work in medical ML has emphasised the importance
of rigorous evaluation protocols, including stratified cross-validation, proper
separation of training and test data, and adherence to reporting guidelines such
as TRIPOD+AI~\cite{Collins2024-tripodai}, to ensure reproducibility and
clinical relevance.

\subsection{Data Augmentation for Imbalanced Tabular Data}

Data augmentation is one of the most widely adopted strategies to address class
imbalance in tabular datasets. Classical oversampling methods, most notably
SMOTE~\cite{Chawla2002-smote} and its variants ADASYN~\cite{He2008-adasyn} and
Borderline-SMOTE~\cite{Han2005-borderline}, generate synthetic samples
through interpolation between minority instances. While effective in certain
binary classification scenarios, these methods were not originally designed for
multiclass problems and apply a uniform interpolation mechanism across all
classes. A growing body of work has highlighted the limitations of such
approaches, particularly in high-dimensional or severely imbalanced settings,
including the introduction of noisy or implausible samples, distortion of class
boundaries, and degradation of model performance when minority classes are
extremely underrepresented~\cite{Blagus2013,Tarawneh2022-multiclass}.

To address these limitations, generative models for tabular data have been
proposed. Among statistically grounded approaches, the Gaussian
Copula~\cite{Patki2016} models multivariate dependencies through a
parametric copula structure, offering a principled alternative to
interpolation. More flexible architectures such as CTGAN and
TVAE~\cite{Xu2019-ctgan} adapt conditional GANs and variational autoencoders
to the mixed numerical--categorical structure of tabular data, and have been
shown to generate more realistic synthetic samples in moderate-data regimes. A
comprehensive review of tabular synthetic data generation methods by Fonseca
and Bacao~\cite{Fonseca2023} surveyed over seventy algorithms and
concluded that deep generative models better capture complex inter-feature
dependencies than interpolation-based approaches, albeit with performance that
varies substantially with dataset size and structural complexity. Similarly,
Sauber-Cole and Khoshgoftaar~\cite{SauberCole2022} provide a focused
survey on the use of GANs to alleviate class imbalance in tabular data,
documenting both their promise and their sensitivity to training
instability in small-sample settings.

Despite these advances, both classical and generative augmentation methods
share a fundamental design assumption: they apply a \emph{uniform} augmentation
strategy across all classes. This assumption is rarely questioned, yet it
ignores the substantial heterogeneity across classes in terms of sample size,
feature distribution, and clinical plausibility. In multiclass medical datasets,
such heterogeneity is the norm rather than the exception, suggesting that
uniform augmentation may be inherently suboptimal. Preliminary observations on
the present dataset are consistent with this concern: uniform treatment
systematically degrades predictive capacity on minority subtypes, either by
introducing implausible synthetic samples (oversampling-based methods) or by
failing to capture stable density estimates when per-class sample size falls
below the regimes for which deep generative models are designed. These
observations motivate a class-dependent augmentation policy in which the choice
of generator is informed by per-class sample size and the resulting
real-to-synthetic ratio.

\subsection{Data Leakage and Experimental Protocols}

In addition to model selection and augmentation design, experimental protocol
plays a critical role in determining the validity of reported results.
Data leakage, defined as the unintended use of information from the test set
during model training, has been identified as one of the most common sources
of performance inflation in applied machine
learning~\cite{Kapoor2023,Arp2024}.

A particularly frequent form of leakage in imbalanced learning arises when data
augmentation is applied prior to train/test splitting, allowing synthetic
samples derived from training instances to appear in the evaluation set. This
violates the independence assumption underlying standard evaluation protocols
and can lead to substantial overestimation of model performance.

Reevaluations of prior work in migraine classification reveal that such issues
are not merely theoretical. In the study by Khan et al.~\cite{Khan2024Migraine},
SMOTE was applied globally before data partitioning, resulting in artificially
inflated accuracy values. When the experimental protocol is corrected by
restricting augmentation to the training folds within stratified
cross-validation, performance drops substantially, and discrepancies between
accuracy and macro-F1 become evident. These findings are consistent with
broader evidence from biomedical ML indicating that leakage and protocol
mis-specification are major drivers of irreproducibility~\cite{Kapoor2023,Arp2024}.

These methodological concerns (majority-dominated metrics, uniform
augmentation strategies, and data leakage) are often discussed in
isolation in the migraine classification literature. The present study
addresses them jointly under a single reproducible protocol, with the aim
of providing a more transparent and conservative assessment of model
performance in this clinical setting.
\section{Methodology}
\label{sec:methodology}

\subsection{Problem Definition and Notation}
\label{subsec:problem_definition}

Let $\mathcal{D} = \{(x_i, y_i)\}_{i=1}^N$ denote a tabular clinical dataset,
where $x_i \in \mathbb{R}^d$ represents a vector of $d$ features describing a
patient, and $y_i \in \mathcal{C} = \{c_1, \dots, c_K\}$ is the corresponding
migraine subtype label. The dataset is characterised by strong class imbalance,
with class frequencies $n_k = |\{i : y_i = c_k\}|$ spanning more than one order
of magnitude.

The goal is to learn a classifier $f: \mathbb{R}^d \rightarrow \mathcal{C}$
that achieves balanced predictive performance across all classes, assigning
equal importance to majority and minority subtypes rather than prioritising
either. Performance is evaluated using macro-averaged F1-score, defined as
the unweighted mean of per-class F1 scores, which operationalises this
balanced objective by giving every class the same weight in the overall
metric~\cite{Powers2011,Saito2015}.

\subsection{Dataset and Experimental Setup}
\label{subsec:dataset}

\begin{figure*}[t]
\centering
\resizebox{\textwidth}{!}{%
\begin{tikzpicture}[
  font=\small,
  >=Latex,
  ampersand replacement=\&,
  box/.style={
    rectangle,
    rounded corners=2pt,
    draw=black!70,
    line width=0.6pt,
    fill=black!3,
    align=center,
    inner sep=4pt,
    minimum height=11mm,
    text width=34mm
  },
  head/.style={
    font=\bfseries\small,
    text=black!85
  },
  arrow/.style={
    -Latex,
    line width=0.7pt,
    draw=black!65
  }
]

\matrix (m) [matrix of nodes,
  nodes={anchor=center},
  column sep=8mm,
  row sep=6mm
]{
  \node[box] (input)
    {Imbalanced Clinical Dataset\\{\footnotesize (400 patients, 7 subtypes,\\14--247 samples per class)}}; \&
  \node[box] (analysis)
    {Per-Class Sample-Size\\Threshold ($\tau = 29$)}; \&
  \node[box, fill=black!6] (small)
    {Small Classes ($n_k < \tau$)\\\textbf{Gaussian Copula}}; \&
  \node[box] (output)
    {Augmented\\Training Set}; \&
  \node[box] (models)
    {Classifier\\{\footnotesize (classical / deep tabular)}}; \&
  \node[box] (eval)
    {Evaluation\\\textbf{macro-F1}\\{\footnotesize (5-fold stratified)}}; \\
  \&
  \&
  \node[box, fill=black!4] (large)
    {Larger Classes ($n_k \geq \tau$)\\\textbf{CTGAN}}; \&
  \\
};

\draw[arrow] (input) -- (analysis);
\draw[arrow] (analysis.east) -- ++(4mm,0) |- (small.west);
\draw[arrow] (analysis.east) -- ++(4mm,0) |- (large.west);
\draw[arrow] (small.east) -- (output.west);
\draw[arrow] (large.east) -- (output.west);
\draw[arrow] (output) -- (models);
\draw[arrow] (models) -- (eval);

\node[head, above=2mm of analysis] {Generator selection};
\node[head, below=4mm of output]
  {\footnotesize Growth mode: \textsc{balance} / proportional $\times 2$ / proportional $\times 4$};

\end{tikzpicture}%
}
\caption{Class-dependent hybrid augmentation framework. The training data
are partitioned by class size: classes with fewer than
$\tau=29$ samples are augmented with Gaussian Copula, while larger
classes are augmented with CTGAN. The volume of synthetic data is
controlled independently by the growth mode (full balance, proportional
$\times 2$, or proportional $\times 4$). The resulting augmented training
set feeds a classifier whose performance is evaluated under stratified
5-fold cross-validation using macro-averaged F1.}
\label{fig:framework}
\end{figure*}
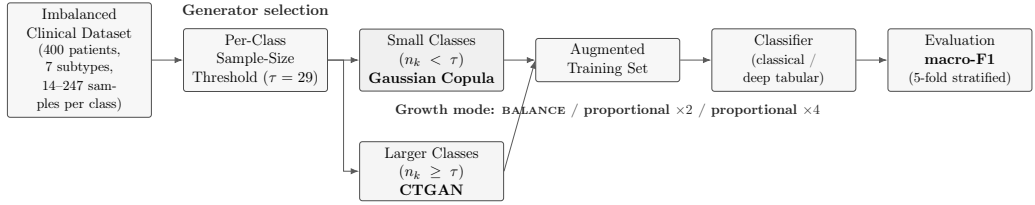

The data used in this study consist of a publicly available migraine
classification corpus that has previously been used by Khan et
al.~\cite{Khan2024Migraine} and Reddy and Reddy~\cite{Reddy2025Migraine};
access details are provided in the Data availability statement at
the end of this paper. According to Khan et al., the underlying
records were originally collected at the Hospital Materno Infantil
de Soledad (Atl\'antico, Colombia). It comprises $N = 400$ patients
described by 22 clinical features after the removal of the
\emph{Ataxia} variable (constant across all patients in the public
release and therefore non-informative). Each record is annotated with
one of seven migraine subtypes defined by the ICHD-3
classification~\cite{IHS2018}, with class sizes ranging from 14 to 247
samples. A second, six-class formulation is additionally evaluated, in
which the two hemiplegic migraine subtypes (sporadic and familial) are
merged into a single ICHD-3-grounded category; the clinical and
statistical justification of this aggregation is presented in
Section~\ref{subsec:discussion_aggregation}. Fig.~\ref{fig:framework}
summarises the resulting class-dependent hybrid augmentation framework,
which is described in detail in
Section~\ref{subsec:hybrid_framework}.

\subsection{Experimental Protocol and Reproducibility}
\label{subsec:experimental_protocol}
All experiments are conducted under stratified 5-fold cross-validation,
ensuring preservation of class distributions across folds. To prevent
data leakage, all preprocessing steps, including data augmentation, are
applied exclusively within the training folds.

Hyperparameter selection is performed via grid search across separate
experimental runs: for each model, candidate configurations are
evaluated under identical fold splits, and the best configuration per
model is carried forward to the final benchmark. To avoid conflating
distributional shift with suboptimal regularisation, hyperparameters
are re-optimised separately for the no-augmentation and augmented
regimes, ensuring that augmented and non-augmented configurations are
compared at their respective optima. The resulting classifier and
augmenter hyperparameters are listed in
Tables~\ref{tab:hyperparams_classifiers}~and~\ref{tab:hyperparams_augmenters}
(\ref{subsec:appendix_hyperparams}). Final performance is
reported as mean $\pm$ standard deviation across the 5 folds.

Augmentation is performed using SDV~\cite{Patki2016} for Gaussian
Copula and CTGAN, and imbalanced-learn~\cite{Lemaitre2017} for the
SMOTE-family oversamplers. Each augmenter is fitted independently
within each training fold and is required to meet a per-augmenter
minimum sample requirement on every minority class; configurations
that would not satisfy this requirement are reported as failures
rather than substituted by an alternative method, preventing silent
fallbacks across augmentation strategies. All runs use a fixed
random seed of~$42$. Library versions, runtime configuration and
hardware are detailed in \ref{subsec:appendix_libraries}
(Table~\ref{tab:libraries}).

\subsection{Exploration of Model and Augmentation Spaces}

Before exploring new models, we reproduced every classifier reported in the
two prior studies under the leakage-free protocol described above. From Khan
et al.~\cite{Khan2024Migraine} we reproduced Support Vector Machine, k-Nearest
Neighbors, Decision Tree, Random Forest and a deep neural network; from Reddy
and Reddy~\cite{Reddy2025Migraine} we reproduced Logistic Regression, Support
Vector Machine, Random Forest and an artificial neural network. Reported
hyperparameters were respected exactly, and unreported ones followed library
defaults (documented per model in \ref{subsec:appendix_sota}). Evaluated under
stratified 5-fold cross-validation with no augmentation, the strongest prior
classifier reaches macro-F1 0.803, with a mean of 0.731 across the eight
configurations and a range of 0.556 to 0.803; the full per-classifier results,
with and without SMOTE, are reported in Table~\ref{tab:sota_replication}
(\ref{subsec:appendix_sota}). We adopt the strongest leakage-free prior model
as the comparison baseline. Starting from this baseline, we performed a
systematic exploration of both the classifier space and the data augmentation
space.

\subsubsection{Classification Models}

The evaluated models include classical machine learning algorithms as Support
Vector Machines (SVM), k-Nearest Neighbors (KNN), Random Forest (RF), XGBoost and deep learning architectures specifically designed for
tabular data.

The latter span four architectural families with distinct inductive biases:
\begin{itemize}
    \item \textbf{MLP}: a feedforward fully connected network applied
    directly to the standardised feature vector, without any
    inter-feature interaction module beyond the linear layers.
    \item \textbf{TabNet}~\cite{Arik2021TabNet}: a sequential architecture
    that applies sparse attention masks at each decision step to perform
    instance-wise feature selection, combining the masks with an additive
    aggregation of step outputs.
    \item \textbf{FT-Transformer}~\cite{Gorishniy2021-fttransformer}:
    tokenises each numerical and categorical feature into a learned
    embedding and processes the resulting sequence with multi-head
    self-attention transformer blocks, enabling explicit modelling of
    pairwise feature interactions.
    \item \textbf{GANDALF}~\cite{Joseph2024GANDALF}: stacks Gated Feature
    Learning Units (GFLU) that apply learnable sparse gating to feature
    representations; rather than computing attention weights over a
    sequence, GFLU implements a multiplicative selection mechanism over
    individual features.
\end{itemize}

Both classical and deep models are included to provide a comprehensive
benchmark across architectural families. The MLP serves as a deep
tabular baseline against which the architectural innovations of the
remaining three models (sparse step-wise attention, full self-attention,
gated feature units) can be measured.

\subsubsection{Data Augmentation Techniques}

We evaluated augmentation techniques spanning three main families:

\paragraph{(i) Oversampling-based methods}
SMOTE and its variants (including Borderline-SMOTE, ADASYN, SVM-SMOTE,
SMOTE-ENN and SMOTE-Tomek) generate synthetic samples through
interpolation between minority-class instances. SMOTE-ENN and
SMOTE-Tomek further combine SMOTE oversampling with edge-cleaning
rules (Edited Nearest Neighbors and Tomek-link removal, respectively)
aimed at reducing the boundary noise introduced by interpolation~\cite{Chawla2002-smote,He2008-adasyn,Han2005-borderline,Lemaitre2017}.

\paragraph{(ii) Probabilistic joint-distribution models}
Gaussian Copula~\cite{Patki2016} fits a parametric copula structure to
the joint distribution of features, providing a principled alternative
to interpolation that remains stable in low-sample regimes.

\paragraph{(iii) Deep generative models}
CTGAN and TVAE~\cite{Xu2019-ctgan} adapt conditional GANs and
variational autoencoders to the mixed numerical--categorical structure
of tabular data, aiming to capture non-linear feature dependencies when
sufficient data are available to fit their adversarial or variational
training objectives.

\subsection{Class-Dependent Hybrid Augmentation Framework}
\label{subsec:hybrid_framework}

\subsubsection{Motivation}

During preliminary experimentation, we observed that different augmentation
methods exhibit heterogeneous performance across classes. In particular,
generative models perform better for moderately sized classes, while
probabilistic models are more stable in extremely low-sample regimes.

This observation suggests that the common practice of applying a uniform
augmentation strategy across all classes is suboptimal.

\subsubsection{Framework Definition}

We define a class-dependent augmentation policy as a mapping:
\[
\mathcal{A}: \mathcal{C} \rightarrow \mathcal{M}
\]
where $\mathcal{M}$ is the set of available augmentation methods.

For each class $c_k$, the selected augmentation method $\mathcal{A}(c_k)$
depends on class-specific characteristics, including sample size $n_k$
and feature distribution $\mathcal{D}_k$. Formally, we treat
$\mathcal{A}$ as a policy that, for each class $c_k$, selects the
generator that best preserves the per-class joint feature distribution
while remaining trainable at the available sample size~$n_k$. In
practice, we approximate this selection through a rule-based partition
with a single sample-size threshold $\tau$, complemented by a discrete
\emph{growth mode} that controls the absolute volume of synthetic data
per class. The growth mode is treated as a configuration parameter
rather than as part of the policy itself.

\subsubsection{Augmentation Policy}

The proposed framework defines two complementary decisions: (i) which
generator $\mathcal{A}(c_k)$ to use per class, and (ii) how much synthetic
data to generate.

\paragraph{Generator selection ($\mathcal{A}$)}
A single sample-size threshold $\tau$ partitions classes into two regimes:
\begin{itemize}
    \item For small classes ($n_k < \tau$): Gaussian Copula, which fits a
    parametric joint distribution that remains stable in low-sample regimes.
    \item For larger classes ($n_k \geq \tau$): CTGAN, which captures
    non-linear feature dependencies when sufficient data are available to
    fit its adversarial training.
\end{itemize}
The threshold is fixed at $\tau = 29$ in our experiments, motivated by
two convergent considerations. Xu et al.~\cite{Xu2019-ctgan} explicitly
identify imbalanced categorical columns as a source of severe mode
collapse in CTGAN, observing that ``imbalanced data also leads to
insufficient training opportunities for minor classes'' and that ``if
the training data are randomly sampled during training, the rows that
fall into the minor category will not be sufficiently represented, thus
the generator may not be trained correctly''. Consistent with this
design constraint, our preliminary runs on this dataset showed CTGAN
training instability for classes with $n_k < 29$ (loss divergence and
degenerate per-class output distributions), whereas Gaussian Copula
remained stable across the same range.

\paragraph{Volume control (growth mode)}
Three growth modes are evaluated, all preserving the per-class generator
assignment defined above:
\begin{itemize}
    \item \textsc{balance}: each minority class is augmented up to the size
    of the majority class, achieving full class balance.
    \item \textsc{proportional}~$\times 2$ and \textsc{proportional}~$\times 4$:
    every class is multiplied by a constant factor, preserving the original
    class proportions while uniformly increasing the absolute volume.
\end{itemize}
The proportional modes are introduced as a means of mitigating fidelity
asymmetry, a phenomenon described in
Section~\ref{subsec:discussion_fidelity}. The choice of growth mode is
treated as a configuration parameter selected per model based on
training-fold performance, and the empirical comparison of the three
modes is presented in Section~\ref{subsec:results_hybrid}.

The class-dependent policy and the chosen growth mode are applied
independently within each training fold, so that no synthetic sample
derived from training data ever leaks into the corresponding validation
fold. Software implementation and library versions are documented in \ref{subsec:appendix_libraries}.

\section{Results}
\label{sec:results}

\subsection{Systematic Exploration of Models and Augmentation Techniques}
\label{subsec:results_exploration}

Starting from the leakage-free prior-art baseline, whose strongest
classifier reaches macro-F1 0.803 (the deep neural network of Khan et al.,
see \ref{subsec:appendix_sota}), we conducted a systematic
experimental exploration over the eight classifiers and nine augmenters
(plus a no-augmentation baseline)
described in Section~\ref{sec:methodology}, covering classical machine
learning, deep tabular models and three augmentation families
(SMOTE-based oversampling, GAN/VAE-based generative models and
parametric joint-distribution approaches).

Consistent patterns emerged across models. CTGAN and Gaussian Copula
produced systematic performance gains for several classifiers,
including RF, MLP, and TabNet. In contrast, SMOTE and its variants
degraded performance for most classifiers in this severely imbalanced
multiclass setting, consistent with known limitations of
interpolation-based oversampling when class boundaries are non-linear or
heavily overlapping~\cite{Saez2016}.

After systematic hyperparameter tuning, the best result in the original
seven-class configuration was obtained by FT-Transformer without
augmentation (macro-F1~$= 0.845 \pm 0.029$) under stratified 5-fold
cross-validation; the complete grid of all eight classifiers against the
ten augmentation conditions is reported in
Tables~\ref{tab:full_pivot_7c_classical}~and~\ref{tab:full_pivot_7c_deep}
(\ref{subsec:appendix_7class}). FT-Transformer thus improves on the best
leakage-free prior model (0.803) by $+0.042$ in the seven-class setting.
Across the entire seven-class benchmark, FT-Transformer without
augmentation is the only configuration to reach 0.845, and no other
model--augmentation combination exceeds 0.82 (GANDALF without
augmentation, 0.817). This concentration of performance just below 0.85,
despite extensive variation in both classifier and augmentation strategy,
points to a structural ceiling attributable to label granularity rather
than insufficient model capacity, as confirmed by the per-class analysis
in the next subsection.

\subsection{Results after Clinically Motivated Class Aggregation}
\label{subsec:results_aggregation}

Given the ceiling observed in the seven-class setting, the classification
task was reformulated into a six-class problem by aggregating two
clinically related subtypes (see
Section~\ref{subsec:discussion_aggregation} for clinical and
methodological justification). This reconfiguration produced a
substantial and consistent improvement across all evaluated models.

Under the six-class configuration, the FT-Transformer without data
augmentation achieved a macro-F1 of $0.896 \pm 0.038$, while GANDALF
reached $0.883 \pm 0.014$ under the same cross-validation protocol.
These results represent a marked improvement over all configurations
evaluated in the original seven-class setting and over the best
leakage-free prior model (0.803, a $+0.093$ gain for FT-Transformer),
and demonstrate that
label granularity constitutes a more binding constraint on
classification performance than augmentation volume in this dataset.

\subsection{Class-Dependent Hybrid Augmentation with Proportional Growth}
\label{subsec:results_hybrid}

Building on these observations, we evaluated the proposed class-dependent
hybrid augmentation framework defined in
Section~\ref{sec:methodology}. For each class, the
generator is assigned according to the sample-size threshold $\tau = 29$
(Gaussian Copula below the threshold, CTGAN above). The volume of
synthetic data is then determined by the growth mode: \textsc{balance}
augments every minority class to the size of the majority class;
\textsc{proportional}~$\times 2$ and \textsc{proportional}~$\times 4$
multiply each class by a constant factor while preserving the original
class proportions.

The proportional modes are designed to mitigate \emph{fidelity asymmetry}
(formalised in Section~\ref{subsec:discussion_fidelity}): in fully
balanced augmentation, minority classes are dominated by synthetic
samples, whereas the majority class remains entirely real, exposing
classifiers to per-class distributions of unequal quality. Proportional
augmentation enforces a uniform fraction of real data across all
classes, providing a more homogeneous learning signal. The choice of
growth mode is treated as a configuration parameter selected per model
based on training-fold performance.

\subsection{Final Benchmark and Best Configurations}
\label{subsec:results_final_benchmark}

Table~\ref{tab:hybrid_vs_individual} consolidates the final
six-class benchmark, comparing the no-augmentation baseline against the
two best individual generators (CTGAN, Gaussian Copula) and the
class-dependent hybrid framework with the best growth mode selected per
classifier.

\begin{table}[ht]
\centering

\caption{
Macro-averaged F1 scores by classifier and augmentation strategy
on the six-class configuration.
Results are reported as mean $\pm$ standard deviation across
stratified 5-fold cross-validation.
The best result per classifier is highlighted in bold.
}

\label{tab:hybrid_vs_individual}

\scriptsize
\setlength{\tabcolsep}{2.5pt}

\begin{tabular}{lcccccc}
\hline

\textbf{Cls.} &
\textbf{No} &
\textbf{CTGAN} &
\textbf{G.C.} &
\textbf{Hyb.} &
\textbf{Hyb.} &
\textbf{Hyb.} \\

&
\textbf{Aug.} &
&
&
\textbf{Bal.} &
$\times 2$ &
$\times 4$ \\

\hline

SVM
& $0.766\pm0.028$
& $0.814\pm0.038$
& $0.823\pm0.047$
& $\mathbf{0.832\pm0.040}$
& $0.819\pm0.061$
& $0.827\pm0.065$ \\

KNN
& $0.795\pm0.077$
& $0.772\pm0.050$
& $0.802\pm0.059$
& $0.825\pm0.043$
& $0.843\pm0.067$
& $\mathbf{0.859\pm0.046}$ \\

RF
& $0.819\pm0.059$
& $0.758\pm0.013$
& $\mathbf{0.848\pm0.021}$
& $0.832\pm0.064$
& $0.840\pm0.073$
& $0.806\pm0.061$ \\

XGBoost
& $0.746\pm0.048$
& $0.799\pm0.020$
& $0.808\pm0.057$
& $0.817\pm0.062$
& $\mathbf{0.823\pm0.050}$
& $0.795\pm0.056$ \\

MLP
& $0.813\pm0.042$
& $0.847\pm0.054$
& $0.857\pm0.052$
& $0.832\pm0.046$
& $\mathbf{0.865\pm0.071}$
& $0.846\pm0.082$ \\

TabNet
& $0.686\pm0.071$
& $0.807\pm0.062$
& $0.839\pm0.048$
& $\mathbf{0.871\pm0.043}$
& $0.775\pm0.072$
& $0.831\pm0.077$ \\

FTT
& $0.896\pm0.038$
& $0.862\pm0.059$
& $0.862\pm0.067$
& $0.854\pm0.060$
& $\mathbf{0.914\pm0.047}$
& $0.867\pm0.042$ \\

GANDALF
& $0.883\pm0.014$
& $0.863\pm0.040$
& $0.850\pm0.042$
& $0.866\pm0.046$
& $0.883\pm0.034$
& $\mathbf{0.895\pm0.039}$ \\

\hline

\textbf{Avg.}
& $0.801$
& $0.815$
& $0.836$
& $0.841$
& $0.845$
& $0.841$ \\

\hline
\end{tabular}

\vspace{1mm}

\parbox{0.98\linewidth}{
\footnotesize
FTT = FT-Transformer.
The hybrid average of $0.862$ reported in the Abstract corresponds
to the per-classifier maximum across the three hybrid growth modes.
}

\end{table}

To complement Table~\ref{tab:hybrid_vs_individual},
Tables~\ref{tab:full_pivot_classical}~and~\ref{tab:full_pivot_deep}
report the complete pivot of all evaluated augmenters against the
classical and deep tabular classifiers respectively, including the
SMOTE family and TVAE excluded from
Table~\ref{tab:hybrid_vs_individual} for clarity. The split by
classifier family follows the architectural distinction introduced in
Section~\ref{sec:methodology} and supports the augmenter-wise
comparisons used in Section~\ref{sec:discussion}.

\begin{table*}[ht]
\centering
\caption{Macro-F1 (mean $\pm$ standard deviation, stratified 5-fold
cross-validation) for the six-class configuration across all evaluated
augmenters and the four \emph{classical} classifiers. The rightmost
column reports the augmenter-wise average across the four classical
classifiers; the bottom row reports the classifier-wise average across
augmenters.}
\label{tab:full_pivot_classical}
\footnotesize
\setlength{\tabcolsep}{5pt}
\begin{tabular}{lccccc}
\hline
\textbf{Augmenter} & \textbf{SVM} & \textbf{KNN} & \textbf{RF} &
\textbf{XGBoost} & \textbf{Avg.} \\
\hline
No augmentation
  & $0.766 \pm 0.028$ & $0.795 \pm 0.077$ & $0.819 \pm 0.059$
  & $0.746 \pm 0.048$ & $0.782$ \\
SMOTE
  & $0.718 \pm 0.009$ & $0.689 \pm 0.076$ & $0.780 \pm 0.044$
  & $0.763 \pm 0.048$ & $0.738$ \\
Borderline-SMOTE
  & $0.711 \pm 0.022$ & $0.708 \pm 0.054$ & $0.748 \pm 0.035$
  & $0.750 \pm 0.026$ & $0.729$ \\
ADASYN
  & $0.701 \pm 0.053$ & $0.697 \pm 0.091$ & $0.769 \pm 0.015$
  & $0.739 \pm 0.023$ & $0.727$ \\
SVM-SMOTE
  & $0.762 \pm 0.057$ & $0.715 \pm 0.056$ & $0.780 \pm 0.033$
  & $0.763 \pm 0.054$ & $0.755$ \\
SMOTE-ENN
  & $0.661 \pm 0.032$ & $0.619 \pm 0.063$ & $0.726 \pm 0.050$
  & $0.712 \pm 0.057$ & $0.680$ \\
SMOTE-Tomek
  & $0.681 \pm 0.025$ & $0.687 \pm 0.067$ & $0.767 \pm 0.036$
  & $0.754 \pm 0.046$ & $0.722$ \\
TVAE
  & $0.716 \pm 0.046$ & $0.733 \pm 0.036$ & $0.766 \pm 0.041$
  & $0.777 \pm 0.050$ & $0.748$ \\
CTGAN
  & $0.814 \pm 0.038$ & $0.772 \pm 0.050$ & $0.758 \pm 0.013$
  & $0.799 \pm 0.020$ & $0.786$ \\
Gaussian Copula
  & $0.823 \pm 0.047$ & $0.802 \pm 0.059$ & $0.848 \pm 0.021$
  & $0.808 \pm 0.057$ & $0.820$ \\
\hline
\textbf{Classifier avg.}
  & $0.735$ & $0.722$ & $0.776$ & $0.761$ & --- \\
\hline
\end{tabular}
\end{table*}

\begin{table*}[ht]
\centering
\caption{Macro-F1 (mean $\pm$ standard deviation, stratified 5-fold
cross-validation) for the six-class configuration across all evaluated
augmenters and the four \emph{deep tabular} classifiers. The rightmost
column reports the augmenter-wise average across the four deep
classifiers; the bottom row reports the classifier-wise average across
augmenters. Abbreviation: FTT = FT-Transformer.}
\label{tab:full_pivot_deep}
\footnotesize
\setlength{\tabcolsep}{5pt}
\begin{tabular}{lccccc}
\hline
\textbf{Augmenter} & \textbf{MLP} & \textbf{TabNet} &
\textbf{FTT} & \textbf{GANDALF} & \textbf{Avg.} \\
\hline
No augmentation
  & $0.813 \pm 0.042$ & $0.686 \pm 0.071$ & $0.896 \pm 0.038$
  & $0.883 \pm 0.014$ & $0.820$ \\
SMOTE
  & $0.727 \pm 0.071$ & $0.737 \pm 0.043$ & $0.774 \pm 0.052$
  & $0.760 \pm 0.063$ & $0.750$ \\
Borderline-SMOTE
  & $0.751 \pm 0.100$ & $0.720 \pm 0.065$ & $0.788 \pm 0.030$
  & $0.767 \pm 0.037$ & $0.757$ \\
ADASYN
  & $0.747 \pm 0.074$ & $0.715 \pm 0.065$ & $0.796 \pm 0.053$
  & $0.748 \pm 0.065$ & $0.752$ \\
SVM-SMOTE
  & $0.766 \pm 0.060$ & $0.708 \pm 0.029$ & $0.830 \pm 0.039$
  & $0.791 \pm 0.009$ & $0.774$ \\
SMOTE-ENN
  & $0.672 \pm 0.094$ & $0.637 \pm 0.061$ & $0.711 \pm 0.050$
  & $0.690 \pm 0.023$ & $0.678$ \\
SMOTE-Tomek
  & $0.670 \pm 0.080$ & $0.731 \pm 0.097$ & $0.766 \pm 0.044$
  & $0.750 \pm 0.061$ & $0.729$ \\
TVAE
  & $0.743 \pm 0.073$ & $0.709 \pm 0.034$ & $0.813 \pm 0.065$
  & $0.795 \pm 0.025$ & $0.765$ \\
CTGAN
  & $0.847 \pm 0.054$ & $0.807 \pm 0.062$ & $0.862 \pm 0.059$
  & $0.863 \pm 0.040$ & $0.845$ \\
Gaussian Copula
  & $0.857 \pm 0.052$ & $0.839 \pm 0.048$ & $0.862 \pm 0.067$
  & $0.850 \pm 0.042$ & $0.852$ \\
\hline
\textbf{Classifier avg.}
  & $0.759$ & $0.729$ & $0.810$ & $0.790$ & --- \\
\hline
\end{tabular}
\end{table*}

The hybrid strategy achieves a higher global average performance
(macro-F1~$= 0.862$) than no augmentation (0.801), pure CTGAN (0.815),
and pure Gaussian Copula (0.836), outperforming individual augmenters in
seven of the eight evaluated classifiers. The single exception is
Random Forest, where Gaussian Copula (0.848) slightly exceeds the best
hybrid configuration (0.840); the difference falls well within the
per-fold standard deviation, consistent with the known sensitivity of
bagged tree ensembles to small distributional shifts in the training
set.

\paragraph{Best validated configuration.}
The strongest result observed across the benchmark is FT-Transformer
under the proportional $\times 2$ growth mode, reaching macro-F1
$= 0.914 \pm 0.047$, a $+0.111$ gain over the best leakage-free prior
model (0.803). The closest non-attention configuration is GANDALF
under proportional $\times 4$, with macro-F1 $= 0.895 \pm 0.039$, a
$+0.012$ improvement over its no-augmentation baseline. Both top
configurations use the re-optimised hyperparameters listed in
Table~\ref{tab:hyperparams_classifiers}
(\ref{subsec:appendix_hyperparams}), tailored to the augmented
regime.

Formal pairwise significance testing across folds is left to future
work. We note that the observed improvements (e.g., $\Delta = +0.012$
for GANDALF~$+$~hybrid$_{\times 4}$, $\Delta = +0.018$ for
FT-Transformer~$+$~hybrid$_{\times 2}$) are of the same order of
magnitude as the per-model fold-level standard deviation
($0.038$--$0.047$). Reported gains should therefore be interpreted as
point estimates rather than as statistically adjudicated effects, and
any claim of ``best configuration'' carries the inherent variance of a
5-fold evaluation on a 400-sample dataset.

\section{Analysis and Discussion}
\label{sec:discussion}

The results presented in Section~\ref{sec:results} support three
principal claims. First, classifier choice and augmentation family
contribute jointly to performance, with neither factor dominating;
both must be tuned together rather than studied in isolation. Second,
uniform augmentation strategies (and SMOTE-based interpolation in
particular) can actively degrade performance in severely imbalanced
multiclass settings. Third, label granularity
constitutes a more binding constraint on classification performance
than augmentation volume, with clinically motivated class aggregation
producing larger gains than any augmentation strategy evaluated. The
proposed class-dependent hybrid framework, which adapts the generator
to per-class sample size and controls volume through a discrete growth
mode, outperforms individual augmenters in seven of the eight evaluated
classifiers and provides the best validated configuration of the study.
The following subsections analyse the mechanisms underlying each of
these findings.

\subsection{Performance Ceiling in the Seven-Class Formulation}
\label{subsec:discussion_ceiling}

The persistent performance ceiling observed in the original seven-class
formulation was not attributable to insufficient model capacity, lack of
hyperparameter tuning, or limited augmentation diversity. Instead,
class-wise analysis (Table~\ref{tab:classwise_f1}, \ref{subsec:appendix_classwise}) revealed a pronounced
bottleneck driven by two subtypes: \emph{Sporadic Hemiplegic Migraine}
(per-class F1~$\approx 0.57$) and \emph{Familial Hemiplegic Migraine}
(F1~$\approx 0.66$), while the remaining five classes achieved a mean
per-class F1 of approximately $0.94$. This asymmetry constrained the
achievable macro-F1, regardless of the classifier or augmentation
strategy employed.

\subsection{Clinically and Statistically Justified Class Aggregation}
\label{subsec:discussion_aggregation}

This bottleneck reflects an intrinsic limitation of the dataset rather than a
modelling failure. According to the ICHD-3~\S1.2.3 classification~\cite{IHS2018},
sporadic and familial hemiplegic migraine correspond to subtypes of the same
hemiplegic migraine syndrome and share identical symptomatic criteria. The sole diagnostic distinction is the presence of a
positive family history, encoded in the dataset by the DPF
(\emph{Direct Positive Family history}) variable, a binary indicator of
whether the patient has at least one first- or second-degree relative
diagnosed with hemiplegic migraine.

While DPF separates the labels, it does not provide discriminative structure
for learning the syndrome itself. Quantitative analysis supports this
interpretation: of the twenty-two features evaluated, eighteen exhibit
absolute mean differences below 0.3 between the two classes, with nine
features taking identical constant values in both subtypes (full
per-feature breakdown in Table~\ref{tab:hemiplegic_feature_diffs}, \ref{subsec:appendix_hemiplegic_diffs}). Clustering analysis
yields near-zero Silhouette coefficients in opposite directions
($+0.123$ for sporadic and $-0.072$ for familial), neither indicating
genuine class cohesion. Consequently, the decision boundary separating
these labels is weakly supported by the remaining feature space and is
largely non-learnable.

Merging both subclasses into a single \emph{Hemiplegic migraine} category is
justified from statistical and machine learning perspectives, and is consistent
with the symptomatic equivalence established in ICHD-3. However, this
aggregation entails a deliberate trade-off: the distinction between familial and
sporadic forms carries clinical relevance for genetic counselling and family
screening. The merged classifier therefore cannot replace subtype-specific
diagnosis in contexts where this distinction is actionable. Its utility is scoped
to settings where the primary objective is syndrome-level identification, and the
aggregation should be treated as a dataset-specific modelling decision rather
than a general clinical recommendation.

\subsection{Why SMOTE-Based Augmentation Degrades Performance}
\label{subsec:discussion_smote}

The consistent degradation observed with SMOTE-based augmentation, documented in
Section~\ref{subsec:results_exploration} and consistent with known limitations
in high-imbalance settings~\cite{Blagus2013, Saez2016}, is interpretable in
terms of the specific geometry of this dataset.

In this dataset, class boundaries are non-linear and partially overlapping, as evidenced by the near-zero Silhouette coefficients reported in Section~\ref{subsec:discussion_aggregation}. Under these conditions, SMOTE-generated samples from minority classes systematically fall within the feature-space regions of adjacent classes, introducing label noise that degrades macro-F1.

In contrast, distribution-learning approaches such as CTGAN and Gaussian Copula
aim to preserve joint feature dependencies and are better aligned with the
structure of clinical tabular data, where interactions are often non-linear and
non-additive. This effect is particularly pronounced under severe imbalance,
where synthetic samples from minority classes may disproportionately contaminate
regions of the feature space associated with other classes, ultimately harming
macro-F1~\cite{DBLP:journals/isci/SaezLSH15}.

\subsection{Fidelity Asymmetry and Proportional Augmentation}
\label{subsec:discussion_fidelity}
\label{subsec:discussion_proportional}

\paragraph{Definition}
Analysis of fully balanced augmented datasets reveals a structural
issue that, to our knowledge, has not been explicitly formalised in the
tabular augmentation literature. Although the balanced dataset used in
this study is 72.9\% synthetic overall, the fraction of real samples is
highly class-dependent: the majority class remains entirely real,
whereas minority classes contain as little as 7--15\% real samples. For
the smallest class ($n = 17$), the synthetic-to-real ratio reaches $\approx 13.5$:1. We
term this phenomenon \emph{fidelity asymmetry} to distinguish it from
the class-frequency imbalance that motivates augmentation.

\paragraph{Mechanism}
While class imbalance is addressed by oversampling, fidelity asymmetry
is \emph{introduced} by it, representing a second-order effect that is
exacerbated by aggressive balancing. This distinction is related to,
but distinct from, the concept of synthetic data quality degradation
documented in low-sample generative
settings~\cite{Fonseca2023,SauberCole2022}: since estimation error is
inherent to any generative process, that error is amplified
disproportionately for minority classes, and in attention-based
architectures that model cross-feature dependencies these discrepancies
can destabilise learned representations precisely for the classes that
require the greatest discriminative accuracy. From this perspective,
augmentation is best treated not as a uniform preprocessing step, but
as a model-selection problem over data-generating processes.

\paragraph{Mitigation through proportional augmentation}
Proportional augmentation directly addresses fidelity asymmetry by
enforcing a uniform fraction of real samples across all classes while
preserving the original class proportions. By avoiding forced full
balance, this strategy maintains the global structure of the data
distribution and provides a more homogeneous learning signal. The gains
observed under proportional augmentation ($+0.012$ macro-F1 for GANDALF
under $\times 4$ and a peak result of $0.914 \pm 0.047$ for
FT-Transformer under $\times 2$, Table~\ref{tab:hybrid_vs_individual})
match or exceed those obtained by optimising the augmentation family
alone: averaging across the eight classifiers in
Tables~\ref{tab:full_pivot_classical}~and~\ref{tab:full_pivot_deep},
the shift from the SMOTE family ($\sim$0.73) to Gaussian Copula
(0.836) yields $\Delta \approx +0.10$, of the same order as the
per-classifier gains attributable to growth mode selection. Managing data fidelity is therefore at least as
consequential as the choice of generative method, and over-aggressive
balancing can be counterproductive in low-data multiclass medical
settings.

\subsection{Clinical Implications}

From a clinical perspective, the proposed framework offers three
concrete contributions to migraine subtype classification under
realistic data constraints. First, by improving macro-averaged
performance on minority subtypes, the approach increases sensitivity to
the under-represented diagnoses that are most likely to be missed in
routine practice and that carry the highest cost of misclassification,
including hemiplegic and basilar-type variants. Second, the explicit
separation between label design (clinically motivated aggregation) and
learning design (class-dependent augmentation) allows clinicians and
methodologists to negotiate trade-offs transparently: a syndrome-level
six-class formulation is appropriate for triage and treatment
selection, while a finer seven-class formulation may be retained when
family-history information is required for genetic counselling. Third,
the leakage-free protocol and per-fold cross-validation design address
key methodological concerns highlighted by TRIPOD+AI reporting
guidance~\cite{Collins2024-tripodai}, in particular items related to
data partitioning and internal validation; full TRIPOD+AI compliance
(including ethical approval, fairness analysis, calibration metrics
and open science statements) remains a target for future iterations
and is a precondition for integration into clinical decision-support
systems. The framework is therefore best understood as
an evaluation and data-handling template rather than as a deployable
diagnostic tool: external validation on independent cohorts remains the
next step before any clinical deployment.

\section{Conclusions}
\label{sec:conclusions}

This study set out to determine whether the high performance reported in prior
migraine classification work reflects genuine predictive capability or
methodological artefact. The evidence presented here supports the latter: data
leakage, majority-dominated metrics, and uniform augmentation strategies
collectively produce inflated estimates that dissolve under rigorous evaluation.
Correcting these artefacts reduces previously inflated performance estimates,
with the strongest leakage-free prior-art classifier, the deep neural network
of Khan et al., reaching macro-F1 0.803 (no augmentation, stratified 5-fold
cross-validation, seven-class setting), well below the inflated accuracies of
up to 99.7\% originally reported. This figure, rather than representing a
failure, defines the true starting point from which meaningful improvement is
possible.

By performing a reproducibility-oriented reevaluation of prior work, we
established this corrected baseline and showed that widely adopted uniform
augmentation strategies, particularly SMOTE-based methods, can degrade
macro-averaged performance in severely imbalanced multiclass settings. In
contrast, generative distribution-learning approaches such as CTGAN and Gaussian
Copula proved more robust, especially when combined with classifiers capable of
modelling non-linear feature interactions.

A key finding of this work is that apparent performance ceilings may reflect
limitations in problem formulation rather than modelling capacity. Through
clinically and statistically justified aggregation of hemiplegic migraine
subclasses in accordance with the ICHD-3 taxonomy, we removed a non-learnable
decision boundary and substantially improved achievable macro-F1. This result
highlights the importance of aligning label structures with underlying clinical
entities when designing machine learning tasks in healthcare.

Building on these insights, we proposed and evaluated a class-dependent
hybrid augmentation framework that adapts augmentation strategies to
class-specific data characteristics. We introduced the concept of
\emph{fidelity asymmetry} to explain why fully balanced augmentation can be
detrimental in low-data regimes, and showed that proportional augmentation,
which preserves class proportions while equalising real-data fractions,
offers a more stable and effective alternative.

This work contributes (i)~a leakage-free per-classifier reevaluation of
every prior-art model (Table~\ref{tab:sota_replication},
\ref{subsec:appendix_sota}), establishing the strongest prior model (macro-F1
0.803) as a corrected baseline that re-frames the previously reported performance on
this dataset, (ii)~a clinically
grounded label aggregation derived from ICHD-3~\S1.2.3 that resolves
the persistent hemiplegic performance ceiling, (iii)~a class-dependent
hybrid augmentation framework that adapts generation strategies to
per-class sample size, and (iv)~the formalisation of \emph{fidelity
asymmetry} as an underexplored second-order effect of class balancing,
together with an empirical demonstration that proportionally
constrained growth can match or exceed full balance for most
classifiers (Table~\ref{tab:hybrid_vs_individual}).

The historical progression of results illustrates the cumulative
methodological and clinical refinement achieved across this study, from an
inflated accuracy of $\sim$0.997 under leakage (with macro-F1 unreported) to
a rigorous macro-F1 of $0.914 \pm 0.047$ under the best validated
configuration (Table~\ref{tab:progression}). Notably, the largest model-controlled gain attributable to a single decision,
from $0.845$ to $0.896$, was produced not by augmentation but by clinically
motivated label redesign, underscoring that problem formulation is the most
consequential modelling decision in this dataset.

\begin{table}[ht]
\centering
\caption{Macro-F1 attained at successive refinement steps of this
study. Each row corresponds to a distinct model--augmentation
configuration evaluated independently; values across rows are not
additive deltas of a single chain. The largest model-controlled
improvement attributable to a single decision is the
$+0.051$ gain from label redesign (FT-Transformer no-augmentation,
7-class $\rightarrow$ 6-class); the augmentation contribution at the
peak is $+0.018$ (FT-Transformer, 6-class, no-aug $0.896 \rightarrow$
hybrid$_{\times 2}$ $0.914$).
Abbreviation: FTT = FT-Transformer.}
\label{tab:progression}

\begin{tabularx}{\linewidth}{l X c}
\hline
\textbf{Refinement step} & \textbf{Configuration} & \textbf{Macro-F1} \\
\hline
Original study (with leakage)
  & DNN + SMOTE applied before split (Khan et al.\ 99.7\% accuracy
    reported); macro-F1 not reported by the authors
  & ---$^\ast$ \\

Best prior-art classifier (leakage-free)
  & DNN (Khan et al.), no augmentation, 7 classes
  & $0.803 \pm 0.077$ \\

Best classifier (7 classes)
  & FTT, no augmentation, 7 classes
  & $0.845 \pm 0.029$ \\

Clinically guided aggregation
  & FTT, 6 classes, no augmentation
  & $0.896 \pm 0.038$ \\

Best validated configuration
  & FTT + {\ttfamily hybrid\_x2}
  & $0.914 \pm 0.047$ \\
\hline
\end{tabularx}

\vspace{0.4em}
\small $^\ast$\,Macro-F1 was not reported in the original studies (only
accuracy). We reproduced all classifiers from both prior studies under a
leakage-free protocol (no augmentation, stratified 5-fold): Khan et al.
(SVM, KNN, Decision Tree, Random Forest, DNN) and Reddy and Reddy (Logistic
Regression, SVM, Random Forest, ANN). The strongest, the DNN of Khan et al.,
reaches macro-F1 $0.803 \pm 0.077$ (best prior-art classifier, second row),
which we use as the honest starting point for subsequent comparisons; the full
per-model results appear in Table~\ref{tab:sota_replication}
(\ref{subsec:appendix_sota}). The deep tabular architectures introduced in
this work (FT-Transformer, GANDALF, TabNet) do not appear in the prior
literature and are therefore excluded from this reference baseline.
\end{table}

This study has several limitations that should inform the interpretation of its
findings. First, the dataset comprises 400 patients from a single clinical
centre, which limits the diversity of clinical phenotypes captured. Second, no
external validation cohort is currently available; generalisation of the
proposed framework to other clinical sites, populations or migraine taxonomies
remains untested and is a precondition for any clinical deployment. Third, the
class-dependent augmentation policy relies on a single sample-size threshold
($\tau$) and a choice of growth mode, both selected for the present dataset;
their transferability to other class distributions has not been assessed.
Fourth, the concept of fidelity asymmetry introduced in
Section~\ref{subsec:discussion_fidelity}, while empirically supported here,
lacks a formal theoretical treatment. Fifth, the performance improvement
associated with class aggregation depends on the specific non-learnability
of the familial, sporadic boundary in this feature set; datasets that
include genetic markers or family history encoded with greater resolution
may not exhibit the same ceiling. Sixth, the variance of the
best-performing configuration (FT-Transformer + hybrid$_{\times 2}$,
$\pm 0.047$, see Section~\ref{subsec:results_final_benchmark}) is
larger than the variance of the no-augmentation FT-Transformer baseline
($\pm 0.038$) and warrants caution in deployment contexts where
predictive stability is as important as mean performance. Finally, all experiments use a fixed
random seed ($42$) for reproducibility, which leaves residual variability
attributable to model initialisation and synthetic generation
unquantified.

\paragraph{Future work}
Several extensions follow naturally from these findings. First,
external validation on independent clinical cohorts is required
before any clinical deployment. Second, multi-seed averaging would
tighten estimates of the variance attributable to model initialisation
and synthetic generation, complementing the fixed-seed protocol used
here. Third, the concept of fidelity asymmetry warrants formal
theoretical treatment, including a characterisation of the conditions
under which it dominates other sources of augmentation-induced
variance. Fourth, a sensitivity analysis on the threshold $\tau$ would
clarify the transferability of the class-dependent policy to other
class distributions. Finally, formal pairwise statistical testing
(Wilcoxon signed-rank with Holm-Bonferroni correction) is recommended
to adjudicate the observed deltas against per-fold variance.

Despite these limitations, the findings have broader methodological implications.
They suggest that data augmentation should be treated as a context-sensitive
design choice rather than a uniform preprocessing step, and that careful
alignment between clinical taxonomy, data fidelity, and evaluation protocol is
essential for trustworthy medical machine learning. The framework proposed
here, by combining leakage-free evaluation, class-dependent augmentation, and
proportionally constrained growth, provides a reproducible and clinically
grounded template that is directly applicable to any multiclass medical
classification problem characterised by low data volume and severe label
imbalance.

More broadly, this work suggests that improving medical machine learning systems may depend less on increasingly complex models and more on rigorous problem formulation, data fidelity, and evaluation design.

\section*{CRediT authorship contribution statement}
\textbf{Elvin Som\'on S\'anchez:} Conceptualization, Methodology, Software,
Investigation, Formal analysis, Data curation, Visualization, Writing --
original draft. \textbf{Miguel A. Guti\'errez-Naranjo:} Conceptualization,
Methodology, Supervision, Writing -- review \& editing.

\section*{Ethics statement}
This study analyses a publicly available migraine classification
dataset hosted on Kaggle that contains no personally identifiable
information. No new patient data were collected for the present
work, and no individual identifiers were processed. 

\section*{Funding}

This work was supported by grant CEX2024-001517-M
(IMUS, ``Apoyo a Unidades de Excelencia Mar\'ia de Maeztu''),
funded by MICIU/\allowbreak AEI/\allowbreak 10.13039/\allowbreak 501100011033. The funding source had
no role in the study design; in the collection, analysis and
interpretation of data; in the writing of the report; or in the
decision to submit the article for publication.

\section*{Declaration of competing interest}
The authors declare that they have no known competing financial
interests or personal relationships that could have appeared to
influence the work reported in this paper.

\section*{Data availability}
The migraine classification dataset analysed in this study is
publicly available on Kaggle\footnote{\url{https://www.kaggle.com/datasets/weinoose/migraine-classification}.}
The same dataset was used in the prior studies by Khan et
al.~\cite{Khan2024Migraine} and Reddy and Reddy~\cite{Reddy2025Migraine}.
No new data were generated for the present work. Experimental
configurations, hyperparameter grids and per-fold cross-validation
artefacts that support the findings of this study are available from
the corresponding author upon reasonable request.

\appendix

\section{Supplementary Tables}
\label{sec:appendix}

This appendix consolidates the implementation details that support the
main benchmark, namely (i)~the exact hyperparameter configurations used
to obtain the macro-F1 values reported in
Tables~\ref{tab:hybrid_vs_individual},~\ref{tab:full_pivot_classical}~and~\ref{tab:full_pivot_deep},
(ii)~the Python libraries and classes underlying each component of the
pipeline, (iii)~the complete macro-F1 grid for the original seven-class
formulation (\ref{subsec:appendix_7class}), (iv)~per-class F1 scores
under both label configurations, and (v)~the leakage-free reevaluation of
every prior-art classifier that defines the comparison baseline
(\ref{subsec:appendix_sota}).

\subsection{Hyperparameter Configurations}
\label{subsec:appendix_hyperparams}

Tables~\ref{tab:hyperparams_classifiers}~and~\ref{tab:hyperparams_augmenters}
report the hyperparameter values applied to each classifier and
augmenter in the final benchmark. Classifier hyperparameters were
selected via grid search on the six-class configuration as described in
Section~\ref{subsec:experimental_protocol}; augmenter hyperparameters
followed library defaults except where noted.

\begin{table*}[ht]
\centering
\caption{Final hyperparameters per classifier (six-class benchmark).
Deep-tabular models additionally used the default early-stopping
configuration of pytorch-tabular~1.2 (monitor:
\texttt{valid\_loss}, patience: $3$, min\_delta: $0.001$, mode:
\texttt{min}) with checkpoint selection on the same metric, and
identical fold splits across all models.
Abbreviation: FTT = FT-Transformer.}
\label{tab:hyperparams_classifiers}
\small
\setlength{\tabcolsep}{4pt}
\begin{tabular}{ll}
\hline
\textbf{Classifier} & \textbf{Hyperparameters} \\
\hline
SVM            & \texttt{C=100}, \texttt{kernel=rbf}, \texttt{gamma=0.1}, \texttt{class\_weight=balanced} \\
KNN            & \texttt{n\_neighbors=5}, \texttt{weights=uniform}, \texttt{metric=euclidean} \\
Random Forest  & \texttt{n\_estimators=300}, \texttt{max\_depth=None}, \texttt{class\_weight=balanced} \\
XGBoost        & \texttt{n\_estimators=300}, \texttt{max\_depth=3}, \texttt{learning\_rate=0.2} \\
MLP            & \texttt{layers=256-128}, \texttt{dropout=0.2}, \texttt{learning\_rate=0.01}, \\
               & \texttt{batch\_size=32}, \texttt{max\_epochs=100} \\
TabNet         & \texttt{n\_d=32}, \texttt{n\_a=32}, \texttt{n\_steps=3}, \texttt{gamma=2.0}, \\
               & \texttt{learning\_rate=0.02}, \texttt{batch\_size=64}, \texttt{max\_epochs=100} \\
FTT            & \texttt{input\_embed\_dim=128}, \texttt{num\_heads=8}, \texttt{num\_attn\_blocks=2}, \\
               & \texttt{attn\_dropout=0.1}, \texttt{ff\_dropout=0.1}, \texttt{learning\_rate=5e-4}, \\
               & \texttt{batch\_size=64}, \texttt{max\_epochs=150} \\
GANDALF        & \texttt{gflu\_stages=10}, \texttt{gflu\_dropout=0.01}, \\
               & \texttt{gflu\_feature\_init\_sparsity=0.1}, \texttt{learning\_rate=5e-3}, \\
               & \texttt{batch\_size=64}, \texttt{max\_epochs=100} \\
\hline
\end{tabular}
\end{table*}

\begin{table*}[ht]
\centering
\caption{Final hyperparameters per augmenter. SMOTE-family
$k\_neighbors$ is adapted at runtime to
$\min(5, n_{\min}-1)$ when the smallest class would otherwise
violate the neighborhood constraint. Hybrid configurations use the
threshold $\tau=29$ described in Section~\ref{subsec:hybrid_framework}.}
\label{tab:hyperparams_augmenters}
\small
\setlength{\tabcolsep}{4pt}
\begin{tabular}{ll}
\hline
\textbf{Augmenter} & \textbf{Hyperparameters} \\
\hline
SMOTE             & \texttt{k\_neighbors=5} (adaptive) \\
Borderline-SMOTE  & \texttt{k\_neighbors=5} (adaptive), \texttt{kind=borderline-1} \\
ADASYN            & \texttt{n\_neighbors=5} (adaptive) \\
SVM-SMOTE         & \texttt{k\_neighbors=5} (adaptive), \texttt{m\_neighbors=10} \\
SMOTE-ENN         & SMOTE \texttt{k\_neighbors=5} + ENN \texttt{n\_neighbors=3} \\
SMOTE-Tomek       & SMOTE \texttt{k\_neighbors=5} + Tomek-link cleaning \\
CTGAN             & \texttt{epochs=300}, \texttt{batch\_size=64}, \texttt{pac=1}, \\
                  & \texttt{generator\_dim=(128,128)} (or $(64,64)$ if $n_k<30$), \\
                  & \texttt{discriminator\_dim} idem \\
TVAE              & \texttt{epochs=300}, \texttt{batch\_size=64}, \\
                  & \texttt{compress\_dims=(64,64)}, \texttt{decompress\_dims=(64,64)} \\
Gaussian Copula   & SDV defaults, \texttt{min\_samples=10} preflight check \\
Hybrid            & $\tau=29$; \texttt{generator(small)}: Gaussian Copula; \\
                  & \texttt{generator(large)}: CTGAN; \\
                  & \texttt{growth\_mode} $\in \{\textsc{balance},\,\textsc{proportional}\,\times 2,$ \\
                  & $\textsc{proportional}\,\times 4\}$ \\
\hline
\end{tabular}
\end{table*}

\subsection{Software Stack and Runtime Environment}
\label{subsec:appendix_libraries}

All experiments were implemented in Python~3.11 within a dedicated
conda environment.
Classical models were trained with scikit-learn~1.4; deep tabular
models with pytorch-tabular~1.2 over PyTorch~2.10 and
PyTorch-Lightning~2.6. Experiments ran on an Apple~M1~Max workstation
(10-core CPU, 32-core integrated GPU, 32\,GB unified memory). Deep
tabular models were trained on the integrated GPU through PyTorch's
native MPS backend; classical models on CPU.
Table~\ref{tab:libraries} maps every component of the experimental
pipeline to the exact Python class used in the implementation.

\begin{table*}[ht]
\centering
\caption{Python libraries and classes used for each pipeline component.
Library version is reported alongside each entry.
Abbreviation: FTT = FT-Transformer.}
\label{tab:libraries}
\small
\setlength{\tabcolsep}{4pt}
\resizebox{\textwidth}{!}{%
\begin{tabular}{lll}
\hline
\textbf{Component} & \textbf{Library (version)} & \textbf{Class} \\
\hline
\multicolumn{3}{l}{\emph{Classical classifiers}} \\
SVM            & scikit-learn~1.4 & \texttt{svm.SVC} \\
KNN            & scikit-learn~1.4 & \texttt{neighbors.KNeighborsClassifier} \\
Random Forest  & scikit-learn~1.4 & \texttt{ensemble.RandomForestClassifier} \\
XGBoost        & xgboost~2.0      & \texttt{XGBClassifier} \\
\hline
\multicolumn{3}{l}{\emph{Deep tabular classifiers (via pytorch-tabular~1.2)}} \\
MLP$^{\ddagger}$ & pytorch-tabular~1.2 & \texttt{CategoryEmbeddingModelConfig} \\
TabNet         & pytorch-tabular~1.2 & \texttt{TabNetModelConfig} \\
FTT            & pytorch-tabular~1.2 & \texttt{FTTransformerConfig} \\
GANDALF        & pytorch-tabular~1.2 & \texttt{GANDALFConfig} \\
\hline
\multicolumn{3}{l}{\emph{Oversampling augmenters}} \\
SMOTE             & imbalanced-learn~0.12 & \texttt{over\_sampling.SMOTE} \\
Borderline-SMOTE  & imbalanced-learn~0.12 & \texttt{over\_sampling.BorderlineSMOTE} \\
ADASYN            & imbalanced-learn~0.12 & \texttt{over\_sampling.ADASYN} \\
SVM-SMOTE         & imbalanced-learn~0.12 & \texttt{over\_sampling.SVMSMOTE} \\
SMOTE-ENN         & imbalanced-learn~0.12 & \texttt{combine.SMOTEENN} \\
SMOTE-Tomek       & imbalanced-learn~0.12 & \texttt{combine.SMOTETomek} \\
\hline
\multicolumn{3}{l}{\emph{Generative and parametric augmenters (via SDV~1.10)}} \\
CTGAN            & SDV~1.10 & \texttt{CTGANSynthesizer} \\
TVAE             & SDV~1.10 & \texttt{TVAESynthesizer} \\
Gaussian Copula  & SDV~1.10 & \texttt{GaussianCopulaSynthesizer} \\
\hline
\multicolumn{3}{l}{\emph{Cross-validation, metrics, framework}} \\
Stratified $k$-fold CV & scikit-learn~1.4 & \texttt{model\_selection.StratifiedKFold} \\
Macro-F1 metric        & scikit-learn~1.4 & \texttt{metrics.f1\_score(average='macro')} \\
Deep-model trainer     & PyTorch-Lightning~2.6 & \texttt{pytorch.Trainer} \\
Tensor backend         & PyTorch~2.10 & \texttt{torch} \\
\hline
\end{tabular}%
}

\vspace{0.4em}
\small $^{\ddagger}$\,The MLP is instantiated through pytorch-tabular's
\texttt{CategoryEmbeddingModelConfig}, but with all 22 features
configured as continuous (\texttt{categorical\_cols=[]}). Under this
configuration no embedding layer is constructed, and the architecture
reduces to a classical multilayer perceptron applied directly to the
standardised feature vector.
\end{table*}

\subsection{Complete Seven-Class Benchmark}
\label{subsec:appendix_7class}

Tables~\ref{tab:full_pivot_7c_classical}~and~\ref{tab:full_pivot_7c_deep}
report the complete macro-F1 grid for the original seven-class
formulation, mirroring the six-class pivots of
Tables~\ref{tab:full_pivot_classical}~and~\ref{tab:full_pivot_deep}. These
results substantiate the performance ceiling discussed in
Section~\ref{subsec:results_exploration}: across all eighty
model--augmentation combinations, FT-Transformer without augmentation is
the single configuration reaching $0.845$, and no other combination
exceeds $0.82$. The corresponding per-class breakdown in
Table~\ref{tab:classwise_f1} attributes this ceiling to the two hemiplegic
subtypes rather than to insufficient model capacity.

\begin{table*}[ht]
\centering
\caption{Macro-F1 (mean $\pm$ standard deviation, stratified 5-fold
cross-validation) for the \emph{seven-class} configuration across all
evaluated augmenters and the four \emph{classical} classifiers. Layout
mirrors Table~\ref{tab:full_pivot_classical} (six-class). The rightmost
column reports the augmenter-wise average across the four classical
classifiers; the bottom row reports the classifier-wise average across
augmenters.}
\label{tab:full_pivot_7c_classical}
\footnotesize
\setlength{\tabcolsep}{5pt}
\begin{tabular}{lccccc}
\hline
\textbf{Augmenter} & \textbf{SVM} & \textbf{KNN} & \textbf{RF} &
\textbf{XGBoost} & \textbf{Avg.} \\
\hline
No augmentation
  & $0.662 \pm 0.075$ & $0.680 \pm 0.038$ & $0.747 \pm 0.087$
  & $0.711 \pm 0.053$ & $0.700$ \\
SMOTE
  & $0.597 \pm 0.046$ & $0.627 \pm 0.085$ & $0.730 \pm 0.043$
  & $0.711 \pm 0.082$ & $0.666$ \\
Borderline-SMOTE
  & $0.601 \pm 0.069$ & $0.639 \pm 0.072$ & $0.703 \pm 0.043$
  & $0.695 \pm 0.089$ & $0.659$ \\
ADASYN
  & $0.615 \pm 0.071$ & $0.622 \pm 0.075$ & $0.698 \pm 0.061$
  & $0.678 \pm 0.087$ & $0.653$ \\
SVM-SMOTE
  & $0.610 \pm 0.055$ & $0.605 \pm 0.069$ & $0.720 \pm 0.053$
  & $0.693 \pm 0.071$ & $0.657$ \\
SMOTE-ENN
  & $0.575 \pm 0.055$ & $0.544 \pm 0.067$ & $0.673 \pm 0.047$
  & $0.637 \pm 0.042$ & $0.607$ \\
SMOTE-Tomek
  & $0.588 \pm 0.052$ & $0.626 \pm 0.085$ & $0.722 \pm 0.053$
  & $0.680 \pm 0.072$ & $0.654$ \\
TVAE
  & $0.624 \pm 0.053$ & $0.628 \pm 0.070$ & $0.710 \pm 0.081$
  & $0.706 \pm 0.085$ & $0.667$ \\
CTGAN
  & $0.744 \pm 0.071$ & $0.688 \pm 0.072$ & $0.756 \pm 0.063$
  & $0.768 \pm 0.069$ & $0.739$ \\
Gaussian Copula
  & $0.728 \pm 0.091$ & $0.705 \pm 0.082$ & $0.748 \pm 0.062$
  & $0.780 \pm 0.063$ & $0.740$ \\
\hline
\textbf{Classifier avg.}
  & $0.634$ & $0.636$ & $0.721$ & $0.706$ & --- \\
\hline
\end{tabular}
\end{table*}

\begin{table*}[ht]
\centering
\caption{Macro-F1 (mean $\pm$ standard deviation, stratified 5-fold
cross-validation) for the \emph{seven-class} configuration across all
evaluated augmenters and the four \emph{deep tabular} classifiers. Layout
mirrors Table~\ref{tab:full_pivot_deep} (six-class). FT-Transformer
without augmentation ($0.845$) is the single highest cell of the entire
seven-class grid. Abbreviation: FTT = FT-Transformer.}
\label{tab:full_pivot_7c_deep}
\footnotesize
\setlength{\tabcolsep}{5pt}
\begin{tabular}{lccccc}
\hline
\textbf{Augmenter} & \textbf{MLP} & \textbf{TabNet} &
\textbf{FTT} & \textbf{GANDALF} & \textbf{Avg.} \\
\hline
No augmentation
  & $0.737 \pm 0.105$ & $0.662 \pm 0.101$ & $0.845 \pm 0.029$
  & $0.817 \pm 0.050$ & $0.765$ \\
SMOTE
  & $0.700 \pm 0.046$ & $0.639 \pm 0.069$ & $0.689 \pm 0.059$
  & $0.730 \pm 0.052$ & $0.690$ \\
Borderline-SMOTE
  & $0.690 \pm 0.065$ & $0.621 \pm 0.062$ & $0.715 \pm 0.039$
  & $0.692 \pm 0.041$ & $0.679$ \\
ADASYN
  & $0.676 \pm 0.028$ & $0.635 \pm 0.081$ & $0.683 \pm 0.029$
  & $0.711 \pm 0.028$ & $0.676$ \\
SVM-SMOTE
  & $0.700 \pm 0.053$ & $0.709 \pm 0.105$ & $0.722 \pm 0.049$
  & $0.759 \pm 0.022$ & $0.722$ \\
SMOTE-ENN
  & $0.626 \pm 0.043$ & $0.565 \pm 0.031$ & $0.681 \pm 0.073$
  & $0.628 \pm 0.051$ & $0.625$ \\
SMOTE-Tomek
  & $0.684 \pm 0.020$ & $0.659 \pm 0.058$ & $0.674 \pm 0.060$
  & $0.693 \pm 0.043$ & $0.677$ \\
TVAE
  & $0.663 \pm 0.037$ & $0.615 \pm 0.093$ & $0.727 \pm 0.075$
  & $0.722 \pm 0.040$ & $0.682$ \\
CTGAN
  & $0.742 \pm 0.064$ & $0.718 \pm 0.038$ & $0.806 \pm 0.034$
  & $0.798 \pm 0.052$ & $0.766$ \\
Gaussian Copula
  & $0.762 \pm 0.033$ & $0.738 \pm 0.042$ & $0.754 \pm 0.060$
  & $0.805 \pm 0.065$ & $0.765$ \\
\hline
\textbf{Classifier avg.}
  & $0.698$ & $0.656$ & $0.730$ & $0.735$ & --- \\
\hline
\end{tabular}

\vspace{0.4em}
\parbox{0.98\linewidth}{\footnotesize
The MLP column denotes the standardised feedforward baseline of
Table~\ref{tab:libraries} (pytorch-tabular
\texttt{CategoryEmbeddingModelConfig}, all features continuous) and is
distinct from the prior-art deep networks of Khan et al.\ (DNN) and Reddy
and Reddy (ANN), which are reproduced under their original configurations
in Table~\ref{tab:sota_replication} (\ref{subsec:appendix_sota}). It is
included as the canonical deep-tabular baseline against which the
attention- and gating-based architectures (TabNet, FT-Transformer,
GANDALF) are measured.}
\end{table*}

\subsection{Per-Class F1 under Both Label Configurations}
\label{subsec:appendix_classwise}

Table~\ref{tab:classwise_f1} reports the per-class F1 scores obtained
by FT-Transformer (no augmentation) under the seven-class and the
six-class formulations, computed as the mean across the five
cross-validation folds. The same model is also reported under the
proportional $\times 2$ hybrid augmentation in the six-class setting,
the configuration that produced the peak macro-F1 of the study
(Section~\ref{subsec:results_final_benchmark}).

\begin{table*}[ht]
\centering
\caption{Per-class F1 (mean $\pm$ std across 5 folds) for FT-Transformer
under two configurations: seven-class no-aug and six-class no-aug.
The hemiplegic row illustrates the effect of clinically motivated
class aggregation: the merged class recovers $\approx +0.07$ F1 over
the 7-class average of its two source subtypes (Sporadic 0.573,
Familial 0.660 $\rightarrow$ merged 0.691). Dashes denote classes
that do not exist under that scheme.}
\label{tab:classwise_f1}
\small
\setlength{\tabcolsep}{4pt}
\begin{tabular}{lcc}
\hline
\textbf{Class} & \textbf{7-class (no aug)} & \textbf{6-class (no aug)} \\
\hline
Migraine without aura
  & $0.968 \pm 0.029$ & $0.967 \pm 0.016$ \\
Typical aura with migraine
  & $0.954 \pm 0.010$ & $0.957 \pm 0.011$ \\
Typical aura without migraine
  & $1.000 \pm 0.000$ & $1.000 \pm 0.000$ \\
Basilar-type aura
  & $0.843 \pm 0.109$ & $0.843 \pm 0.109$ \\
Other
  & $0.920 \pm 0.098$ & $0.920 \pm 0.098$ \\
Sporadic hemiplegic migraine
  & $0.573 \pm 0.142$ & --- \\
Familial hemiplegic migraine
  & $0.660 \pm 0.095$ & --- \\
Hemiplegic migraine (merged)
  & --- & $0.691 \pm 0.094$ \\
\hline
\textbf{Macro-F1}
  & $0.845 \pm 0.029$ & $0.896 \pm 0.038$ \\
\hline
\end{tabular}
\end{table*}

Table~\ref{tab:classwise_f1} shows that all five non-hemiplegic
classes already operate near their per-class ceiling under the
seven-class formulation; the macro-F1 improvement from 0.845 to
0.896 in the six-class setting is therefore attributable almost
entirely to the collapse of the two hemiplegic subtypes into a
single, learnable class.

\subsection{Per-Feature Separability between Sporadic and Familial
Hemiplegic Migraine}
\label{subsec:appendix_hemiplegic_diffs}

Table~\ref{tab:hemiplegic_feature_diffs} reports, for each of the 22
clinical features (after Ataxia removal), the per-class mean for
Sporadic ($n=14$) and Familial ($n=24$) Hemiplegic Migraine, together
with the absolute difference $|\bar{x}_S - \bar{x}_F|$. The DPF feature
is shown for completeness as the only perfect separator (binary
family-history indicator). Of the twenty-two features, eighteen exhibit
absolute mean differences below $0.3$, including nine features that take
identical constant values in both subtypes. The remaining four
non-trivial separators (Intensity, Sensory, Visual, DPF) account for
the residual class signal, of which only DPF is perfectly informative.
These figures support the symptomatic equivalence between the two
subtypes claimed in Section~\ref{subsec:discussion_aggregation}.

\begin{table}[ht]
\centering
\caption{Per-feature absolute mean differences between Sporadic
($n=14$) and Familial ($n=24$) Hemiplegic Migraine, sorted by
$|\bar{x}_S - \bar{x}_F|$ ascending. DPF$^\dagger$ is the only perfect
separator (encoded as $0$ for Sporadic and $1$ for Familial). All
remaining features are encoded as small non-negative integers in the
public dataset.}
\label{tab:hemiplegic_feature_diffs}
\small
\setlength{\tabcolsep}{4pt}
\begin{tabular}{lccc}
\hline
\textbf{Feature} & $\bar{x}_S$ & $\bar{x}_F$ & $|\Delta|$ \\
\hline
Photophobia & 1.000 & 1.000 & 0.000 \\
Location    & 1.000 & 1.000 & 0.000 \\
Character   & 1.000 & 1.000 & 0.000 \\
Diplopia    & 0.000 & 0.000 & 0.000 \\
Nausea      & 1.000 & 1.000 & 0.000 \\
Hypoacusis  & 0.000 & 0.000 & 0.000 \\
Phonophobia & 1.000 & 1.000 & 0.000 \\
Defect      & 0.000 & 0.000 & 0.000 \\
Paresthesia & 0.000 & 0.000 & 0.000 \\
Tinnitus    & 0.286 & 0.292 & 0.006 \\
Frequency   & 1.643 & 1.667 & 0.024 \\
Vertigo     & 0.357 & 0.417 & 0.060 \\
Dysarthria  & 0.071 & 0.000 & 0.071 \\
Duration    & 1.500 & 1.583 & 0.083 \\
Conscience  & 0.000 & 0.083 & 0.083 \\
Age         & 21.571 & 21.458 & 0.113 \\
Dysphasia   & 0.429 & 0.292 & 0.137 \\
Vomit       & 0.429 & 0.208 & 0.220 \\
\hline
Intensity   & 2.143 & 2.500 & 0.357 \\
Sensory     & 0.571 & 0.208 & 0.363 \\
Visual      & 1.786 & 1.333 & 0.452 \\
DPF$^\dagger$ & 0.000 & 1.000 & 1.000 \\
\hline
\end{tabular}

\vspace{0.4em}
\small $^\dagger$\,DPF (\emph{Direct Positive Family history}) is a
binary indicator and the only feature that perfectly separates the two
subtypes by construction; it does not provide discriminative structure
for the underlying syndrome.
\end{table}

\subsection{Replication of prior studies classifiers under
leakage-free evaluation}
\label{subsec:appendix_sota}

To establish an honest comparison baseline, we reproduced every classifier
reported by Khan et al.~\cite{Khan2024Migraine} and Reddy and
Reddy~\cite{Reddy2025Migraine} under the project's leakage-free protocol:
stratified 5-fold cross-validation over the seven ICHD-3 classes, with
augmentation (when used) fitted exclusively within each training fold.
Reported hyperparameters were respected exactly; unreported ones follow
library defaults, documented per model in the note below. Khan et al. name
scikit-learn explicitly and describe the deep network in Keras terms,
whereas Reddy and Reddy report no software and specify few hyperparameters. The Support Vector Machine (linear, $C=1$) is
common to both papers and is therefore computed once.

Two caveats apply to the comparison. First, the original studies reported
\emph{accuracy} (Khan) or classification accuracy / CA (Reddy) obtained
\emph{with} pre-split SMOTE, that is, under data leakage, whereas the values
in Table~\ref{tab:sota_replication} are \emph{macro-F1} obtained \emph{without}
leakage; the two sets of figures are therefore not directly comparable. Under
leakage, Khan et al. reported accuracies between 88\% and 99.7\% (with
augmentation, SVM 94.60\%, KNN 97.10\%, Decision Tree 88.20\%, Random Forest
88.50\%, DNN 99.66\%), and Reddy and Reddy reported selective-sampling CA
between 0.958 and 0.995 (ANN 0.995, Random Forest 0.980, SVM 0.965, Logistic
Regression 0.958). Under the leakage-free protocol the strongest prior model
reaches only macro-F1 0.803, with a mean of 0.731 across the eight
configurations and a range of 0.556 to 0.803.

\begin{table}[ht]
\centering
\caption{Leakage-free replication of the classifiers from the two prior
studies: macro-F1 (mean $\pm$ standard
deviation, stratified 5-fold cross-validation, seven ICHD-3 classes) for
every prior-art classifier reproduced under the leakage-free protocol, with
no augmentation and with SMOTE fitted within each training fold. The best
value per column is shown in bold. These figures are not comparable to the
accuracy / CA values reported in the original studies, which were obtained
with pre-split SMOTE (leakage).}
\label{tab:sota_replication}
\footnotesize
\setlength{\tabcolsep}{5pt}
\begin{tabular}{llcc}
\hline
\textbf{Classifier} & \textbf{Source} & \textbf{No augmentation} &
\textbf{SMOTE} \\
\hline
SVM (linear, $C=1$) & Khan + Reddy & $0.784 \pm 0.062$ & $0.681 \pm 0.040$ \\
KNN                 & Khan         & $0.710 \pm 0.046$ & $0.635 \pm 0.075$ \\
Decision Tree       & Khan         & $0.556 \pm 0.066$ & $0.610 \pm 0.068$ \\
Random Forest (100 trees) & Khan   & $0.739 \pm 0.096$ & $\mathbf{0.729 \pm 0.024}$ \\
DNN                 & Khan         & $\mathbf{0.803 \pm 0.077}$ & $0.702 \pm 0.060$ \\
Logistic Regression & Reddy        & $0.785 \pm 0.064$ & $0.706 \pm 0.028$ \\
Random Forest (10 trees) & Reddy   & $0.671 \pm 0.082$ & $0.700 \pm 0.050$ \\
ANN                 & Reddy        & $0.801 \pm 0.065$ & $0.716 \pm 0.038$ \\
\hline
\end{tabular}

\vspace{0.4em}
\parbox{0.98\linewidth}{\footnotesize
The SVM (linear, $C=1$) is common to both papers (Khan reports it explicitly;
Reddy reports a linear kernel with $C$ unreported, set to $1$) and is computed
once. Random Forest is reported by Khan (scikit-learn defaults, 100 trees) and
by Reddy (a ``minimal number of trees with growth control'', interpreted here
as 10 trees), and is therefore listed separately by source. Settings reported
by the authors were respected. The original studies reported accuracy between
88\% and 99.7\% (Khan) and CA between 0.958 and 0.995 (Reddy) under pre-split SMOTE
(leakage); the macro-F1 values above are the corresponding leakage-free
reevaluations.}
\end{table}

\clearpage

\bibliographystyle{elsarticle-num}
\bibliography{references}

@article{SauberCole2022,
  author  = {Sauber-Cole, Rick and Khoshgoftaar, Taghi M.},
  title   = {The use of generative adversarial networks to alleviate class
             imbalance in tabular data: a survey},
  journal = {Journal of Big Data},
  volume  = {9},
  number  = {1},
  pages   = {98},
  year    = {2022},
  doi     = {10.1186/s40537-022-00648-6}
}

@article{Fonseca2023,
  author  = {Fonseca, Joao and Bacao, Fernando},
  title   = {Tabular and latent space synthetic data generation: a literature review},
  journal = {Journal of Big Data},
  volume  = {10},
  number  = {1},
  pages   = {115},
  year    = {2023},
  doi     = {10.1186/s40537-023-00792-7}
}

@article{DBLP:journals/isci/SaezLSH15,
  author       = {Jos{\'{e}} A. S{\'{a}}ez and
                  Juli{\'{a}}n Luengo and
                  Jerzy Stefanowski and
                  Francisco Herrera},
  title        = {{SMOTE-IPF:} Addressing the noisy and borderline examples problem
                  in imbalanced classification by a re-sampling method with filtering},
  journal      = {Inf. Sci.},
  volume       = {291},
  pages        = {184--203},
  year         = {2015},
  url          = {https://doi.org/10.1016/j.ins.2014.08.051},
  doi          = {10.1016/J.INS.2014.08.051},
  bibsource    = {dblp computer science bibliography, https://dblp.org}
}

@INPROCEEDINGS{Patki2016,
  author={Patki, Neha and Wedge, Roy and Veeramachaneni, Kalyan},
  booktitle={2016 IEEE International Conference on Data Science and Advanced Analytics (DSAA)}, 
  title={The Synthetic Data Vault}, 
  year={2016},
  volume={},
  number={},
  pages={399-410},
  doi={10.1109/DSAA.2016.49}}

@article{Hollmann2022TabPFN,
  author  = {Hollmann, Noah and M{\"u}ller, Samuel and Eggensperger, Katharina and Lindauer, Marius},
  title   = {Tabular data: Deep learning is not all you need},
  journal = {Advances in Neural Information Processing Systems},
  year    = {2022},
  volume  = {35},
  pages   = {644--658}
}

@article{Ashina2021Migraine,
  author = {Messoud Ashina},
  title = {Migraine},
  journal = {The New England Journal of Medicine},
  year = {2021},
  volume = {383},
  number = {19},
  pages = {1866--1876},
  doi = {10.1056/NEJMra1915327}
}

@article{Saez2016,
  author = {Jos{\'e} A. S{\'a}ez and Juli{\'a}n Luengo and Francisco Herrera},
  title = {Evaluating the classifier behavior with noisy data considering performance and robustness},
  journal = {Information Sciences},
  year = {2016},
  volume = {346--347},
  pages = {256--274},
  doi = {10.1016/j.ins.2016.03.050}
}

@article{Collins2024-tripodai,
  author  = {Collins, Gary S. and Moons, Karel G. M. and Dhiman, Paula 
             and Riley, Richard D. and Beam, Andrew L. and 
             Van Calster, Ben and Ghassemi, Marzyeh and Liu, Xiaoxuan 
             and Reitsma, Johannes B. and van Smeden, Maarten and others},
  title   = {{TRIPOD+AI} statement: updated guidance for reporting 
             clinical prediction models that use regression or 
             machine learning methods},
  journal = {BMJ},
  volume  = {385},
  pages   = {e078378},
  year    = {2024},
  doi     = {10.1136/bmj-2023-078378}
}

@article{Borisov2022survey,
  author  = {Borisov, Vadim and Leemann, Tobias and Se{\ss}ler, Kathrin 
             and Haug, Johannes and Pawelczyk, Martin and Kasneci, Gjergji},
  title   = {Deep Neural Networks and Tabular Data: A Survey},
  journal = {IEEE Transactions on Neural Networks and Learning Systems},
  volume  = {35},
  pages   = {7499--7519},
  year    = {2022},
  doi     = {10.1109/TNNLS.2022.3229161}
}

@article{Sokolova2009,
  author  = {Sokolova, Marina and Lapalme, Guy},
  title   = {A systematic analysis of performance measures 
             for classification tasks},
  journal = {Information Processing \& Management},
  volume  = {45},
  number  = {4},
  pages   = {427--437},
  year    = {2009},
  doi     = {10.1016/j.ipm.2009.03.002}
}

@article{ShwartzZiv2022,
  author  = {Shwartz-Ziv, Ravid and Armon, Amitai},
  title   = {Tabular data: Deep learning is not all you need},
  journal = {Information Fusion},
  year    = {2022},
  volume  = {81},
  pages   = {84--90},
  doi     = {10.1016/j.inffus.2021.11.011}
}

@article{Grinsztajn2022,
  author  = {Grinsztajn, Leo and Oyallon, Edouard and Varoquaux, Ga{\"e}l},
  title   = {Why do tree-based models still outperform deep learning on tabular data?},
  journal = {Advances in Neural Information Processing Systems},
  year    = {2022},
  volume  = {35},
  pages   = {507--520}
}

@article{Petrusic2024-recommendations,
  author  = {Petru{\v{s}}i{\'c}, Igor and Savi{\'c}, Andrej 
             and Mitrovi{\'c}, Katarina and Ba{\v{c}}anin, Neboj{\v{s}}a
             and Sebastianelli, Gabriele and Secci, Daniele 
             and Coppola, Gianluca},
  title   = {Machine learning classification meets migraine: 
             recommendations for study evaluation},
  journal = {The Journal of Headache and Pain},
  volume  = {25},
  number  = {1},
  pages   = {215},
  year    = {2024},
  doi     = {10.1186/s10194-024-01924-x}
}

@inproceedings{Somepalli2021SAINT,
  author = {Gautam Somepalli and Micah Goldblum and Avi Schwarzschild and Micah Bruss and Tom Goldstein},
  title = {SAINT: Improved Neural Networks for Tabular Data via Row Attention and Contrastive Pre-Training},
  booktitle = {Advances in Neural Information Processing Systems},
  year = {2021},
  volume = {34},
  pages = {23983--23994}
}

@misc{Joseph2024GANDALF,
  author = {Manu Joseph and Harsh Raj},
  title = {GANDALF: Gated Adaptive Network for Deep Automated Learning of Features},
  year = {2024},
  eprint = {2207.08548},
  archivePrefix = {arXiv},
  doi = {10.48550/arXiv.2207.08548}
}

@article{IHS2018,
  author  = {{International Headache Society}},
  title   = {The International Classification of Headache Disorders, 3rd edition (ICHD-3)},
  journal = {Cephalalgia},
  year    = {2018},
  volume  = {38},
  pages   = {1--211},
  doi     = {10.1177/0333102417738202}
}

@inproceedings{Gorishniy2021-fttransformer,
  author    = {Gorishniy, Yury and Rubachev, Ivan and Khrulkov, Valentin and Babenko, Artem},
  title     = {Revisiting Deep Learning Models for Tabular Data},
  booktitle = {Advances in Neural Information Processing Systems},
  volume    = {34},
  pages     = {18598--18608},
  year      = {2021},
  publisher = {Curran Associates, Inc.},
  url       = {https://proceedings.neurips.cc/paper/2021/hash/9d86d83f925f2149e9edb0ac3b49229c-Abstract.html}
}

@ARTICLE{Saito2015,
  title     = "The precision-recall plot is more informative than the {ROC}
               plot when evaluating binary classifiers on imbalanced datasets",
  author    = "Saito, Takaya and Rehmsmeier, Marc",
  journal   = "PLoS One",
  publisher = "Public Library of Science (PLoS)",
  volume    =  10,
  number    =  3,
  pages     = "e0118432",
  month     =  mar,
  year      =  2015,
  copyright = "http://creativecommons.org/licenses/by/4.0/",
  language  = "en"
}

@ARTICLE{Blagus2013,
  title     = "{SMOTE} for high-dimensional class-imbalanced data",
  author    = "Blagus, Rok and Lusa, Lara",
  journal   = "BMC Bioinformatics",
  publisher = "Springer Science and Business Media LLC",
  volume    =  14,
  number    =  1,
  pages     = "106",
  month     =  mar,
  year      =  2013,
  language  = "en"
}

@ARTICLE{Kapoor2023,
  title     = "Leakage and the reproducibility crisis in machine-learning-based
               science",
  author    = "Kapoor, Sayash and Narayanan, Arvind",
  journal   = "Patterns (N. Y.)",
  publisher = "Elsevier BV",
  volume    =  4,
  number    =  9,
  pages     = "100804",
  month     =  sep,
  year      =  2023,
  copyright = "http://creativecommons.org/licenses/by/4.0/",
  language  = "en"
}

@ARTICLE{Arp2024,
  title     = "Pitfalls in machine learning for computer security",
  author    = "Arp, Daniel and Quiring, Erwin and Pendlebury, Feargus and
               Warnecke, Alexander and Pierazzi, Fabio and Wressnegger,
               Christian and Cavallaro, Lorenzo and Rieck, Konrad",
  journal   = "Commun. ACM",
  publisher = "Association for Computing Machinery (ACM)",
  volume    =  67,
  number    =  11,
  pages     = "104--112",
  month     =  nov,
  year      =  2024,
  copyright = "https://creativecommons.org/licenses/by-nc-nd/4.0/",
  language  = "en"
}

@ARTICLE{Tarawneh2022-multiclass,
  title     = "Stop oversampling for class imbalance learning: A review",
  author    = "Tarawneh, Ahmad S and Hassanat, Ahmad B and Altarawneh, Ghada
               Awad and Almuhaimeed, Abdullah",
  journal   = "IEEE Access",
  publisher = "Institute of Electrical and Electronics Engineers (IEEE)",
  volume    =  10,
  pages     = "47643--47660",
  year      =  2022,
  copyright = "https://creativecommons.org/licenses/by/4.0/legalcode"
}

@article{Xu2019-ctgan,
  author       = {Lei Xu and
                  Maria Skoularidou and
                  Alfredo Cuesta{-}Infante and
                  Kalyan Veeramachaneni},
  title        = {Modeling Tabular data using Conditional {GAN}},
  journal      = {CoRR},
  volume       = {abs/1907.00503},
  year         = {2019},
  url          = {http://arxiv.org/abs/1907.00503},
  eprinttype   = {arXiv},
  eprint       = {1907.00503},
  bibsource    = {dblp computer science bibliography, https://dblp.org}
}

@INCOLLECTION{Han2005-borderline,
  title     = "{Borderline-SMOTE}: A new over-sampling method in imbalanced
               data sets learning",
  booktitle = "Lecture Notes in Computer Science",
  author    = "Han, Hui and Wang, Wen-Yuan and Mao, Bing-Huan",
  publisher = "Springer Berlin Heidelberg",
  pages     = "878--887",
  series    = "Lecture Notes in Computer Science",
  year      =  2005,
  address   = "Berlin, Heidelberg"
}

@INPROCEEDINGS{He2008-adasyn,
  title           = "{ADASYN}: Adaptive synthetic sampling approach for
                     imbalanced learning",
  booktitle       = "2008 {IEEE} International Joint Conference on Neural
                     Networks ({IEEE} World Congress on Computational
                     Intelligence)",
  author          = "He, Haibo and Bai, Yang and Garcia, Edwardo A and Li,
                     Shutao",
  publisher       = "IEEE",
  month           =  jun,
  year            =  2008,
  pages           = "1322--1328",
  conference      = "2008 IEEE International Joint Conference on Neural
                     Networks (IJCNN 2008 - Hong Kong)",
  location        = "Hong Kong, China"
}

@ARTICLE{Chawla2002-smote,
  title     = "{SMOTE}: Synthetic minority over-sampling technique",
  author    = "Chawla, N V and Bowyer, K W and Hall, L O and Kegelmeyer, W P",
  journal   = "J. Artif. Intell. Res.",
  publisher = "AI Access Foundation",
  volume    =  16,
  pages     = "321--357",
  month     =  jun,
  year      =  2002
}

@article{Petrusic2025-ig,
  author  = {Petru{\v{s}}i{\'c}, Igor and Messina, Roberta 
             and Pellesi, Lanfranco and others},
  title   = {Application of machine learning in migraine classification: 
             a call for study design standardization and global 
             collaboration},
  journal = {The Journal of Headache and Pain},
  volume  = {26},
  number  = {1},
  pages   = {200},
  year    = {2025},
  doi     = {10.1186/s10194-025-02134-9}
}

@ARTICLE{Lee2025-jb,
  title     = "The current role of artificial intelligence in the field of
               headache disorders, with a focus on migraine: A systemic review",
  author    = "Lee, Wonwoo and Chu, Min Kyung",
  journal   = "Headache and Pain Research",
  publisher = "The Korean Headache Society",
  month     =  feb,
  year      =  2025,
  language  = "en"
}

@ARTICLE{Stubberud2024-ou,
  title     = "Artificial intelligence and headache",
  author    = "Stubberud, Anker and Langseth, Helge and Nachev, Parashkev and
               Matharu, Manjit S and Tronvik, Erling",
  journal   = "Cephalalgia",
  publisher = "SAGE Publications",
  volume    =  44,
  number    =  8,
  pages     = "3331024241268290",
  month     =  aug,
  year      =  2024,
  language  = "en"
}

@ARTICLE{Mosquera2024-lj,
  title     = "Class imbalance on medical image classification: towards better
               evaluation practices for discrimination and calibration
               performance",
  author    = "Mosquera, Candelaria and Ferrer, Luciana and Milone, Diego H and
               Luna, Daniel and Ferrante, Enzo",
  journal   = "Eur. Radiol.",
  publisher = "Springer Science and Business Media LLC",
  volume    =  34,
  number    =  12,
  pages     = "7895--7903",
  month     =  dec,
  year      =  2024,
  copyright = "https://www.springernature.com/gp/researchers/text-and-data-mining",
  language  = "en"
}

@ARTICLE{Hellin2024-qy,
  title     = "Unraveling the impact of class imbalance on deep-learning models
               for medical image classification",
  author    = "Hell{\'\i}n, Carlos J and Olmedo, Alvaro A and Valledor,
               Adri{\'a}n and G{\'o}mez, Josefa and L{\'o}pez-Ben{\'\i}tez,
               Miguel and Tayebi, Abdelhamid",
  journal   = "Appl. Sci. (Basel)",
  publisher = "MDPI AG",
  volume    =  14,
  number    =  8,
  pages     = "3419",
  month     =  apr,
  year      =  2024,
  copyright = "https://creativecommons.org/licenses/by/4.0/",
  language  = "en"
}

@ARTICLE{He_Garcia_2009,
  author={He, Haibo and Garcia, Edwardo A.},
  journal={IEEE Transactions on Knowledge and Data Engineering}, 
  title={Learning from Imbalanced Data}, 
  year={2009},
  volume={21},
  number={9},
  pages={1263-1284},
  doi={10.1109/TKDE.2008.239}}

@ARTICLE{Stovner2022-az,
  title     = "The global prevalence of headache: an update, with analysis of
               the influences of methodological factors on prevalence estimates",
  author    = "Stovner, Lars Jacob and Hagen, Knut and Linde, Mattias and
               Steiner, Timothy J",
  journal   = "J. Headache Pain",
  publisher = "Springer Science and Business Media LLC",
  volume    =  23,
  number    =  1,
  pages     = "34",
  month     =  apr,
  year      =  2022,
  copyright = "https://creativecommons.org/licenses/by/4.0",
  language  = "en"
}

@article{Khan2024Migraine,
  title={Migraine headache ({MH}) classification using machine learning methods with data augmentation},
  author={Khan, Lal and Shahreen, Moudasra and Qazi, Atika and Shah, Syed Jamil Ahmed and Hussain, Sabir and Chang, Hsien-Tsung},
  journal={Scientific Reports},
  volume={14},
  number={1},
  pages={5180},
  year={2024},
  publisher={Nature Publishing Group},
  doi={10.1038/s41598-024-55874-0}
}

@article{Arik2021TabNet,
  author  = {Arik, Sercan and Pfister, Tomas},
  title   = {TabNet: Attentive Interpretable Tabular Learning},
  journal = {Proceedings of the AAAI Conference on Artificial Intelligence},
  year    = {2021},
  volume  = {35},
  pages   = {6679--6687}
}

@ARTICLE{Reddy2025Migraine,
  title     = "Migraine triggers, phases, and classification using machine
               learning models",
  author    = "Reddy, Anusha and Reddy, Ajit",
  journal   = "Front. Neurol.",
  publisher = "Frontiers Media SA",
  volume    =  16,
  pages     = "1555215",
  month     =  may,
  year      =  2025,
  copyright = "https://creativecommons.org/licenses/by/4.0/",
  language  = "en"
}

@article{Powers2011,
  title={Evaluation: From precision, recall and F-measure to ROC, informedness, markedness and correlation},
  author={Powers, David M.W.},
  journal={Journal of Machine Learning Technologies},
  volume={2},
  number={1},
  pages={37--63},
  year={2011},
  publisher={BioInfo Publications}
}

@article{Lemaitre2017,
  author  = {Lema{\^i}tre, Guillaume and Nogueira, Fernando and Aridas, Christos K.},
  title   = {Imbalanced-learn: A Python Toolbox to Tackle the Curse of Imbalanced Datasets in Machine Learning},
  journal = {Journal of Machine Learning Research},
  volume  = {18},
  number  = {17},
  pages   = {1--5},
  year    = {2017}
}

\end{document}